\documentclass[]{fairmeta}
\microtypesetup{expansion=false}

\usepackage[utf8]{inputenc}
\usepackage{url,array,makecell,colortbl,amsfonts,amsmath,amssymb}
\usepackage{algorithm,algorithmic,newfloat,listings,paracol,titletoc,enumitem,chngcntr,tabularx}

\DeclareCaptionStyle{ruled}{labelfont=normalfont,labelsep=colon,strut=off}
\floatstyle{ruled}
\newfloat{listing}{tb}{lst}{}
\floatname{listing}{Listing}
\newtcblisting[auto counter,number within=section]{spatialcliprompt}[2][]{
  enhanced standard,
  breakable,
  height fixed for=first and middle,
  listing only,
  colback=blue!5!white,
  colframe=blue!75!black,
  title={Box~\thetcbcounter: #2},
  fonttitle=\bfseries,
  boxrule=0.8pt,
  arc=2mm,
  left=1mm,
  right=1mm,
  top=1mm,
  bottom=1mm,
  listing options={
    basicstyle=\footnotesize\ttfamily,
    breaklines=true,
    breakatwhitespace=true,
    breakindent=0pt,
    breakautoindent=false,
    columns=fullflexible,
    keepspaces=true,
    showstringspaces=false,
    frame=none,
    aboveskip=0pt,
    belowskip=0pt
  },
  #1
}
\newtcolorbox{casestep}[2]{
  enhanced standard,
  breakable,
  height fixed for=first and middle,
  colback=#1!5!white,
  colframe=#1!65!black,
  title={#2},
  fonttitle=\bfseries\footnotesize,
  fontupper=\scriptsize,
  boxrule=0.7pt,
  arc=1.5mm,
  left=1.2mm,
  right=1.2mm,
  top=0.8mm,
  bottom=0.8mm,
  before skip=3pt,
  after skip=3pt
}

\title{Deferred Exposure of Future Trajectories for Verifiable Reasoning in Autonomous Driving VLMs}
\author[1,2,\dagger]{Zixuan Huang}
\author[2,3,\dagger]{Yang Zhou}
\author[2]{Kaixuan Wang}
\author[2]{Guli Zhang}
\author[1]{Hongyan Xie}
\author[4]{Yakun Zhu}
\author[1]{Hao Geng}
\author[2]{Xiaozhi Chen}
\author[1,*]{Yikun Ban}
\author[1,*]{Deqing Wang}

\renewcommand{\affiliationlist}{%
\affiliationformat[1]{Beihang University},
\affiliationformat[2]{Zhuoyu Technology, Shenzhen, China},
\affiliationformat[3]{Zhejiang University},
\affiliationformat[4]{Shanghai Jiao Tong University}%
}

\contribution[\dagger]{Equal contribution}
\contribution[*]{Corresponding author}
\date{2026.7.31}
\metadata[Code]{\url{https://github.com/hzx122/DEFT-RLVR}}
\metadata[Model]{\url{https://huggingface.co/hzxllll/DEFT-RLVR-model-HF}}
\metadata[Dataset]{\url{https://huggingface.co/datasets/hzxllll/AD-MCQ}}
\correspondence{\email{huang\_zx@buaa.edu.cn}}

\definecolor{modelblue}{RGB}{230,242,255}
\definecolor{modelpink}{RGB}{255,235,242}
\definecolor{jeftpink}{RGB}{231,161,176}
\definecolor{deftmcqblue}{RGB}{118,169,220}
\definecolor{deftgenorange}{RGB}{233,161,91}
\definecolor{deftrlvrred}{RGB}{214,107,114}
\definecolor{deftsharedgray}{RGB}{120,120,120}

\newcounter{deftalgorithm}
\renewcommand{\thedeftalgorithm}{\arabic{deftalgorithm}}
\newcommand{\deftvariantbox}[2]{%
  \begingroup
  \setlength{\fboxsep}{3pt}%
  \noindent\fcolorbox{#1}{#1!7}{%
    \parbox{\dimexpr\linewidth-2\fboxsep-2\fboxrule\relax}{#2}}%
  \endgroup}

\abstract{Recent Vision-Language-Action (VLA) models for autonomous driving (AD) increasingly utilize chain-of-thought (CoT) supervision to enhance the reasoning capabilities of their Vision-Language Model (VLM) components, yet existing annotation pipelines commonly expose the teacher model to the logged ground-truth (GT) future trajectory.
We empirically show that this induces \textbf{trajectory anchoring bias}: teacher models rationalize the revealed outcome rather than infer a decision from scene evidence, producing less causally faithful CoTs and substantially more severe hallucinations, especially in causally challenging scenes.
Removing the GT trajectory eliminates this shortcut, but open-ended trajectory generation entangles high-level decision-making with precise geometric synthesis and low-level dynamics.
To make trajectory-level driving decisions verifiable without requiring open-ended trajectory synthesis, we introduce \textbf{Autonomous-Driving Multiple-Choice Question (AD-MCQ)}, which casts planning as selection among explicit trajectory candidates.
Taking this a step further, we propose \textbf{Deferred Exposure of Future Trajectories for RLVR (DEFT-RLVR)} to transform future trajectories from pre-decision anchors into post-decision verification targets.
Experimental results show that \textbf{DEFT-RLVR} improves AD reasoning while preserving or even enhancing general visual capabilities.
With VLM-only inference and controllable difficulty through candidate construction, \textbf{AD-MCQ} provides a flexible, scalable, and extensible foundation for future research on verifiable AD reasoning.
}

\begin{document}
\maketitle

\section{Introduction}

Mainstream Vision-Language-Action (VLA) models for autonomous driving (AD) typically couple a large Vision-Language Model (VLM) with a substantially smaller action expert \cite{tian2024drivevlm,jiang2024senna}.
The action expert is typically specialized for geometric prediction, whereas high-level reasoning and decision making fall to the VLM, making its AD-specific reasoning capability critical to downstream planning.

Recent work seeks to enhance this ability through CoT supervision~\cite{zhao2025cot,zhou2026autovla,tian2024drivevlm,wang2024drivecot,gu2026accelerating}.
However, when reasoning must resolve into a concrete driving decision~\cite{zhou2026autovla}, rather than scene understanding alone~\cite{ishaq2025drivelmm,wei2025ad,marcu2024lingoqa,park2025nuplanqa}, 
its CoT supervision is typically conditioned on the ground-truth (GT) trajectory.
\textbf{Given the logged future trajectory, the VLM CoT annotator rationalizes a known outcome rather than inferring a decision from scene evidence.}
This mirrors \textbf{anchoring bias} in cognitive psychology, whereby initially supplied information can disproportionately shape subsequent judgments \cite{yasseri2022fooled}.

\begin{figure}[t]
\centering
\includegraphics[width=\columnwidth]{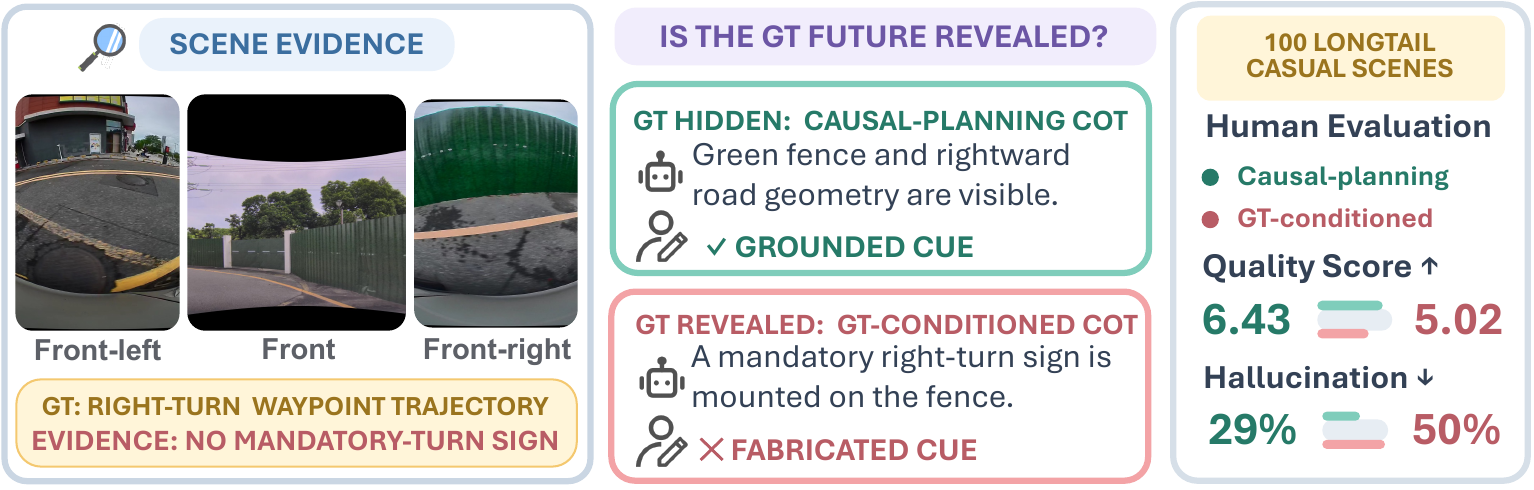}
\caption{GT conditioning induces post-hoc rationalization. The illustrated CoT invents a mandatory-turn sign absent from the scene; aggregate results show lower causal faithfulness and preference, with more severe hallucinations.}
\label{fig:motivation-overview}
\end{figure}

We empirically validate \textbf{trajectory anchoring bias} through the controlled study in Figure~\ref{fig:motivation-overview}.
GT-conditioned CoTs exhibit lower causal faithfulness than causal-planning CoTs,
with the degradation primarily concentrated in hard causal scenarios where reliable reasoning is most critical.
Moreover, exposing the model to the GT trajectory substantially increases the incidence of severe hallucinations,
indicating that trajectory conditioning can inject fabricated causal evidence into the CoT supervision used for subsequent training.

Given this, a natural remedy is to hide the GT trajectory while the teacher derives both its rationale and driving decision from the observed scene, and to verify the predicted future only afterward. This restores the solve-then-verify paradigm used in reasoning-model distillation \cite{shao2024deepseekmath,yang2025qwen3}, rather than revealing the answer before constructing its rationale.

For AD, however, open-ended trajectory synthesis is poorly matched to this paradigm because it entangles high-level decision making with precise continuous geometry and low-level dynamics. 
We therefore seek a language-model-compatible interface through which AD reasoning can emerge from the VLM's general reasoning capability.

To make trajectory-level driving decisions verifiable without open-ended geometric synthesis, we introduce \textbf{Autonomous-Driving Multiple-Choice Question (AD-MCQ)}, a candidate-trajectory benchmark that casts AD planning as selection among scene-specific explicit trajectories. Unlike coarse meta actions, its candidates preserve distinctions in braking time, speed profile, and lateral geometry, grounding each answer in a concrete explicit plan.

\textbf{AD-MCQ} makes trajectory-level driving decisions verifiable, 
but revealing candidate trajectories before reasoning can simply replace the GT-trajectory anchor with a candidate-set anchor: 
the policy model inevitably focuses on comparing the relative quality of trajectories, thereby taking shortcuts in reasoning.
We therefore propose \textbf{Deferred Exposure of Future Trajectories for RLVR (DEFT-RLVR)}, which first commits the policy to a scene-derived decision and only then reveals candidates for explicit grounding.
\textbf{In this way, trajectories supervise reasoning as post-decision targets rather than pre-decision premises.}

Across multiple VLM backbones, \textbf{DEFT-RLVR} consistently strengthens autonomous-driving reasoning and decision making while slightly improving aggregate general visual capability.

In summary, our contributions in this work are as follows:
\begin{itemize}
    \item We identify and empirically validate  \textbf{trajectory anchoring bias}: exposing the demonstrated future trajectory produces action-consistent but causally unfaithful rationales, especially in hard causal scenes.
    \item We introduce \textbf{AD-MCQ}, a verifiable candidate-trajectory benchmark that preserves explicit trajectory-level distinctions and supports both exact selection and candidate-blind reasoning evaluation without open-ended coordinate generation.
    \item We propose \textbf{DEFT-RLVR}, which defers candidate-trajectory exposure until after the policy has committed to a scene-derived decision and uses exact trajectory correctness and  question-specific process supervision. This design improves AD reasoning while preserving the general visual capability of the base policy.
\end{itemize}

\section{Trajectory Anchoring Bias in AD VLMs}
\label{sec:motivation}

Our motivation begins with a simple research question: \emph{does revealing the GT future trajectory help a teacher infer a faithful driving rationale,
or merely make an already known outcome easier to justify?} 
The AD-VLM must infer the appropriate trajectory from the evidence available in the historical scene.
By contrast, the GT trajectory can act as an anchor, allowing teachers to reverse inference and construct a post hoc explanation of the revealed outcome~\cite{xu-etal-2024-preemptive,arcuschin2025chain}.
Consequently, the resulting CoT may be geometrically consistent with the GT trajectory while failing to faithfully identify the scene evidence that genuinely supports the action \cite{balasubramanian-etal-2025-closer,xu2026more}.

\begin{figure}[!h]
\centering
\includegraphics[width=0.6\linewidth]{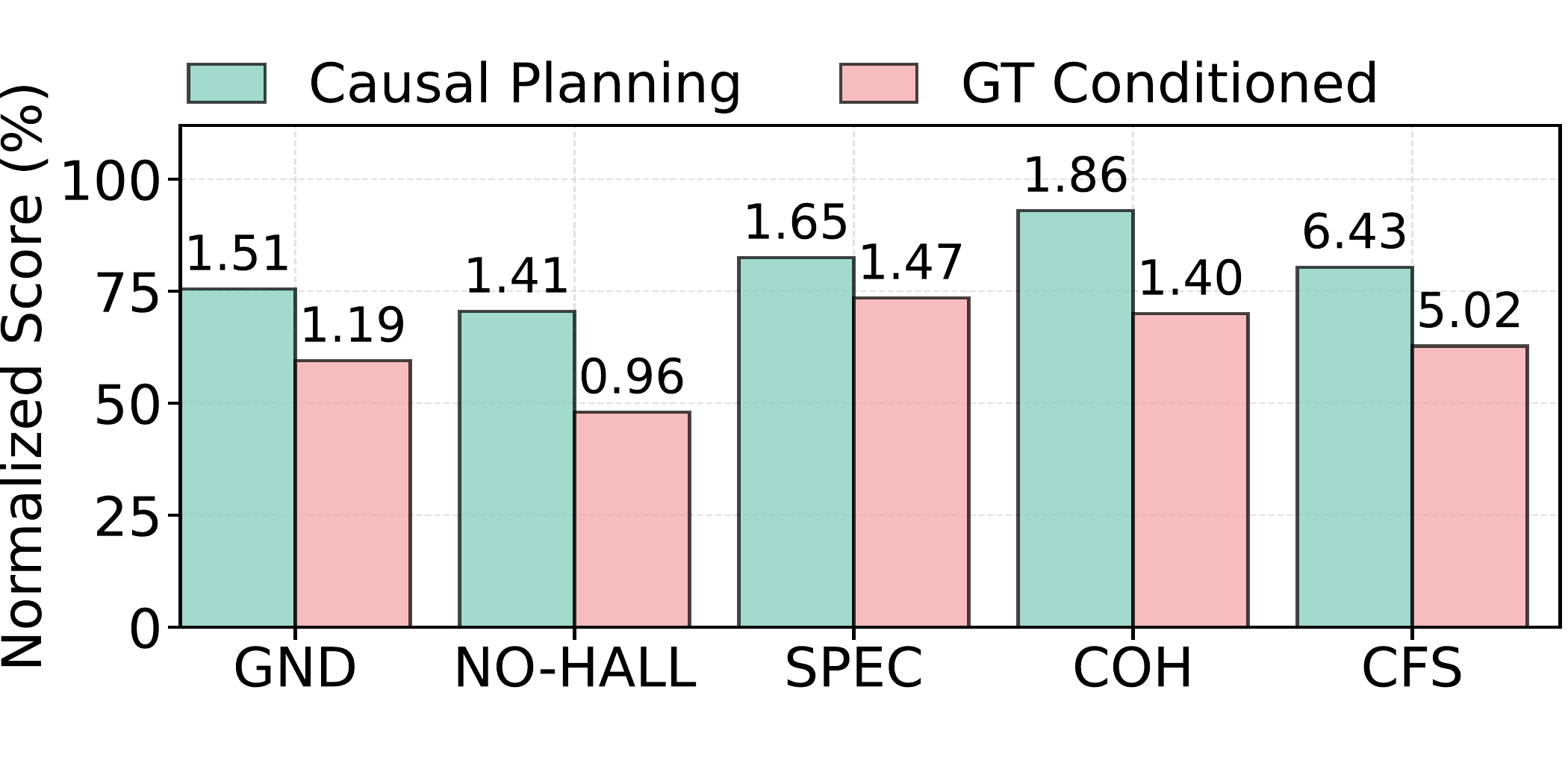}
\caption{Human-rated causal faithfulness comparison.
GT-conditioned CoTs receive lower scores than causal-planning CoTs across grounding (GND), absence of hallucination (NO-HALL), specificity (SPEC), causal coherence (COH), and aggregate causal-faithfulness score (CFS).}
\label{fig:cfs-metric-bars}
\end{figure}

We examine this anchoring hypothesis through a human-scored study summarized in Figure~\ref{fig:cfs-metric-bars} and Table~\ref{tab:motivation-cfs}.
GT-conditioned CoTs exhibit lower causal faithfulness,
a substantially higher incidence of severe hallucination,
and lower pairwise preference than causal-planning CoTs.
The study design and detailed analysis are provided in Appendix~\ref{app:motivation-study}.

\begin{table}
\centering
\begin{tabular}{lcc}
\hline
GT Exposure & Severe Halluc. $\downarrow$ & Pairwise Win $\uparrow$ \\
\hline
No (Causal Planning) & 29.0\% & 60.5\% \\
Yes (GT-Conditioned) & 50.0\% & 24.0\% \\
\hline
\end{tabular}
\caption{Human-rated effect of pre-reasoning GT-trajectory exposure on CoT quality. Exposing the trajectories increases severe hallucinations and reduces pairwise preference.}
\label{tab:motivation-cfs}
\end{table}

These results expose a severe supervision-direction mismatch:
\textbf{for post hoc chain-of-thought annotation of trajectory decisions, access to the future trajectory serves as a reasoning shortcut rather than a decision target.}
We therefore retain trajectories as verifiable targets while excluding them from the premises of causal reasoning: 
\textbf{the model must first infer a plan from the scene and only then ground it in an explicit future.
}

\begin{figure*}[t]
\centering
\includegraphics[width=1.0\linewidth]{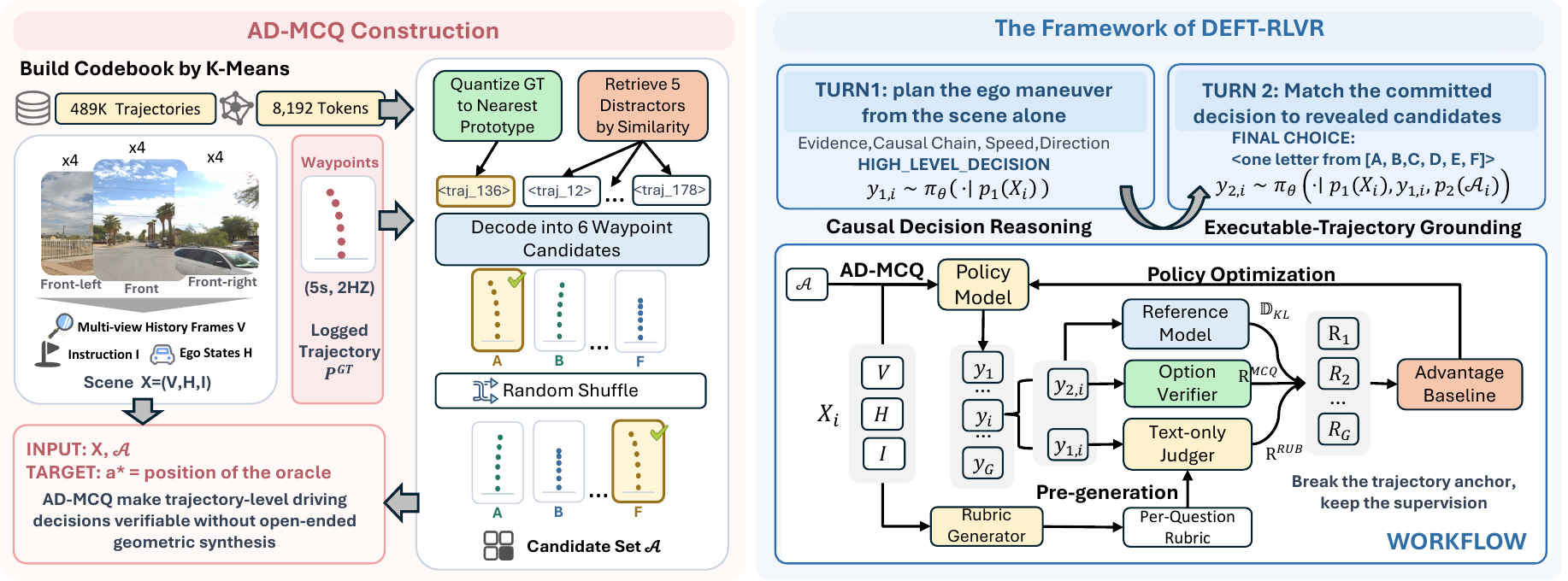}
\caption{The framework of AD-MCQ and DEFT-RLVR. AD-MCQ turns explicit trajectory selection into an exactly verifiable decision; DEFT-RLVR defers candidate exposure and combines outcome correctness with rubric supervision of candidate-blind reasoning.}
\label{fig:rubric-pipeline}
\end{figure*}

\section{\textbf{AD-MCQ}: A Verifiable Candidate-Trajectory Benchmark}
\label{sec:mcq}


As shown in Figure~\ref{fig:rubric-pipeline}, \textbf{AD-MCQ} formulates autonomous-driving planning as selecting a future trajectory from a small, scene-specific candidate set. This formulation preserves trajectory-level granularity while replacing open-ended coordinate generation with an exactly verifiable decision.

\paragraph{Task Formulation.}
Each \textbf{AD-MCQ} instance consists of multi-view scene-history frames $V_i$,
ego history and current motion state $H_i$,
a navigation instruction $I_i$,
and a shuffled set of explicit candidate trajectories
$\mathcal A_i=(\mathbf P_{i,1},\ldots,\mathbf P_{i,M})$.
Exactly one candidate corresponds to the quantized logged future,
and its shuffled position $a_i^\star\in\{1,\ldots,M\}$ serves as the exactly verifiable target.

\paragraph{Discrete Trajectory Prototypes.}
Let a fixed-horizon ego trajectory be
$\mathbf{P}=(\mathbf{p}_1,\ldots,\mathbf{p}_T)\in\mathbb{R}^{T\times 2}$,
where $\mathbf p_t=(x_t,y_t)$ denotes longitudinal and lateral displacement from the current ego pose.
We flatten each trajectory in a corpus of $N$ logged futures and apply $K$-means to obtain a codebook
$\mathcal C=\{\mathbf C_1,\ldots,\mathbf C_K\}$,
where each prototype $\mathbf C_k=(\mathbf c_{k,1},\ldots,\mathbf c_{k,T})$ represents a complete future motion.
We quantize $\mathbf P$ by nearest-prototype assignment:
\begin{equation}
    z(\mathbf{P})=\arg\min_{k\in\{1,\ldots,K\}}
    \sum_{t=1}^{T}\|\mathbf{p}_t-\mathbf{c}_{k,t}\|_2^2.
    \label{eq:traj-token-quantization}
\end{equation}


\begin{figure}[h]
\centering
\includegraphics[width=0.6\columnwidth]{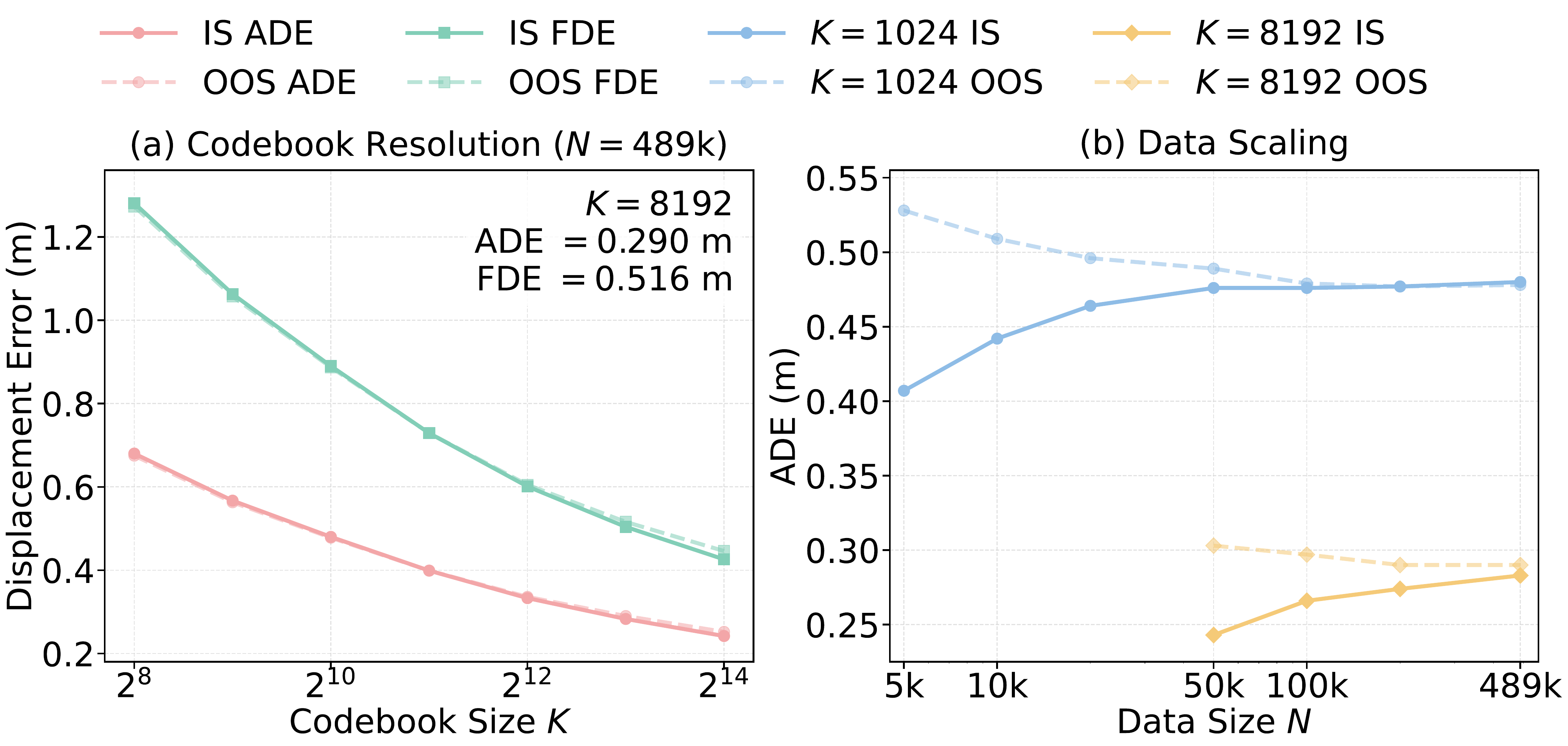}
\caption{Trajectory-codebook scaling. Reconstruction error versus (a) codebook size \(K\) and (b) clustering-corpus size \(N\); solid/dashed curves indicate in-/out-of-sample trajectories.}
\label{fig:traj-token-representation}
\end{figure}

To determine an appropriate codebook configuration, we study how the number of
clustering trajectories \(N\) and prototypes \(K\) affect reconstruction fidelity
and codebook utilization.
Figure~\ref{fig:traj-token-representation} shows that  \(K{=}8192\) provides a favorable balance between out-of-sample reconstruction fidelity and codebook utilization. We provide full construction and scaling analyses in Appendix~\ref{app:traj-token-repr}.

\paragraph{Candidate-trajectory Construction.}
We measure the distance between two prototypes by
\begin{equation}
    d_{ij}=\frac{1}{T}\sum_{t=1}^{T}
    \|\mathbf{c}_{i,t}-\mathbf{c}_{j,t}\|_2,
    \quad
    \rho_{ij}=1-\frac{d_{ij}-d_{\min}}{d_{\max}-d_{\min}},
    \label{eq:traj-token-similarity}
\end{equation}
where larger $\rho_{ij}$ indicates more similar decoded trajectories.
For each driving scene $i$,
we first map the GT trajectory $\mathbf{P}_i^{\mathrm{gt}}$ to its nearest codebook entry
$z_i^\star=z(\mathbf{P}_i^{\mathrm{gt}})$.
Based on Eq.~\ref{eq:traj-token-similarity},
we define the hard-negative pool as
\begin{equation}
    \mathcal{H}(z_i^\star)=
    \left\{z\in[K]\setminus\{z_i^\star\}:
    \rho_{\min}\leq \rho_{z,z_i^\star}\leq \rho_{\max}\right\}.
    \label{eq:mcq-hard-pool}
\end{equation}
We construct split-specific distractors from $\mathcal H(z_i^\star)$ as detailed in Appendix~\ref{app:experimental-details}
and summarized algorithmically in Appendix~\ref{app:ad-mcq-construction},
combine \(M-1\) distinct negatives with $z_i^\star$,
and randomly shuffle the $M$ candidates.
Finally, we decode the candidate indices $(z_{i,1},\ldots,z_{i,M})$ into the waypoint-option set shown to the model:
\begin{equation}
\mathcal A_i
:=(\mathbf{P}_{i,m})_{m=1}^{M}
=\bigl(\mathbf C_{z_{i,m}}\bigr)_{m=1}^{M}
\in(\mathbb{R}^{T\times2})^M.
    \label{eq:mcq-option-set}
\end{equation}

The VLM then selects one shuffled option,
and its decision is evaluated by the deterministic verifier.
Complete instances are provided in Appendix~\ref{app:mcq-case-study}.

Notably, the codebook ultimately retrieves textual waypoint trajectory candidates \cite{li2025discrete}.
Unlike direct waypoint retrieval, the codebook maps continuous futures to a finite motion vocabulary~\cite{philion2023trajeglish,wu2024smart,pertsch2025fast}, 
enabling controlled hard-negative construction without sacrificing explicit geometry.

\section{\textbf{DEFT-RLVR}: Deferred Exposure of Future Trajectories}
\label{sec:method}


Building on \textbf{AD-MCQ}, \textbf{DEFT-RLVR} mitigates trajectory-induced anchoring bias by deferring candidate exposure until after the policy commits to a scene-derived decision, while jointly optimizing the two interaction stages with rubric-based reasoning supervision.

\subsection{DEFT: Deferred Exposure of Future Trajectories}
\label{sec:method-interface}


By deferring candidate exposure, DEFT reserves candidate geometry for grounding an already formed scene-derived decision rather than shaping the decision itself.

Specifically, for question $i$, let $X_i=(V_i,H_i,I_i)$ denote the scene context and
$\mathcal A_i=(\mathbf{P}_{i,m})_{m=1}^{M}$ the candidate trajectories with option label set $\mathcal L_i$.

\textbf{Turn 1: Causal Decision Reasoning.}
Conditioned solely on the scene context $X_i$, the causal-reasoning prompt $p_1$ (provided in Appendix~\ref{app:two-turn-policy-prompt})
elicits causal reasoning over  scene evidence before the candidate trajectories are revealed:
\begin{equation}
u_{1,i}=p_1(X_i),\quad
y_{1,i}\sim\pi_\theta(\cdot\mid u_{1,i}).
\label{eq:turn-one-policy}
\end{equation}
The resulting response \(y_{1,i}\)  explicitly traces how scene evidence leads to driving implications and commits to a complete high-level decision (HLD) before candidate exposure.

\textbf{Turn 2: Explicit-Trajectory Grounding.}
Only after this decision has been formed do we reveal \(\mathcal A_i\) through the trajectory-matching prompt \(p_2\) (provided in Appendix~\ref{app:two-turn-policy-prompt}), which treats the recorded decision as binding and uses candidate geometry only to identify the closest explicit realization of that decision.
With $u_{2,i}=p_2(\mathcal A_i)$, we sample:
\begin{equation}
y_{2,i}\sim\pi_\theta(\cdot\mid u_{1,i},y_{1,i},u_{2,i}).
\label{eq:turn-two-policy}
\end{equation}
The selected option is $\widehat a_i=\operatorname{parse}(y_{2,i})\in\mathcal L_i\cup\{\bot\}$,
where $\bot$ denotes an invalid output.
\textbf{This design prevents candidate geometry from conditioning the initial reasoning process while retaining exact trajectory-level verification.}

\subsection{Joint Optimization of the Two-Stage Interaction}

Although \textbf{DEFT} separates candidate-free decision formation from trajectory grounding, we optimize them jointly as a single rollout using Group Relative Policy Optimization (GRPO)~\cite{guo2025deepseek}.
For each question $i$, we sample $G$ two-turn rollouts under the interaction defined in Section~\ref{sec:method-interface}.
We serialize rollout $j$ as the complete two-turn sequence
$s_{i,j}=u_{1,i}\oplus y_{1,i,j}\oplus u_{2,i}\oplus y_{2,i,j}$.
During optimization, we perform a single forward pass over $s_{i,j}$ to compute the token likelihoods used for importance sampling,
while masking the prompt tokens so that the policy objective is applied only to the generated tokens in $y_{1,i,j}$ and $y_{2,i,j}$.
Both turns share the rollout reward $R_{i,j}$ and its group-normalized advantage.

\subsection{Structured Rubric Rewards for Reasoning-Trace Supervision}
\label{sec:method-rubric}

AD-MCQ provides a verifiable outcome reward:
\begin{equation}
R_{i,j}^{\mathrm{MCQ}}
=\mathbb I[\widehat a_{i,j}=a_i^\star],
\label{eq:rlvr-reward}
\end{equation}
where $a_i^\star$ is the oracle option,
but this signal alone cannot distinguish grounded reasoning from rationalization.

To prevent reinforcing reasoning trajectories that arrive at the correct answer through shortcut exploitation or random guessing \cite{guo2026rethinking}, 
we apply rubric-based reasoning rewards to rollouts with correct MCQ answers.
For each answer-correct rollout,
we form the normalized Turn-1 reasoning trace
\(\widetilde y_{1,i,j}:=\operatorname{Normalize}(y_{1,i,j})\)
and asynchronously submit \(\widetilde y_{1,i,j}\) to a text grader for
evaluation.
Specifically, \textbf{DEFT-RLVR} generates an instance-specific rubric once offline using a vision-language rubric generator $\mathcal G$ conditioned on the fixed generation prompt $p_{\mathrm{rub}}$ (provided in Appendix~\ref{app:offline-rubric-generation-prompt}):
\begin{equation}
\mathcal C_i
=\mathcal G\!\left( p_{\mathrm{rub}} (X_i)\right).
\label{eq:offline-rubric-generation}
\end{equation}


The resulting rubric \(\mathcal C_i=\{(c_{i,k},w_{i,k})\}_{k=1}^{K_i}\) contains atomic, positively weighted criteria that explicitly encode scene evidence verified by the offline generator and is reused across rollouts~\cite{gunjal2025rubrics,rezaei2025online,rao2026autorubric}.

During RL rollouts, we prompt a shared VLM judge $\mathcal J$ with $p_\mathrm{txt}$ (provided in Appendix~\ref{app:online-text-grader-prompt}) to evaluate each reasoning trace against the rubric criteria set $\mathcal{C}_i$:
\begin{equation}
\mathbf b_{i,j}=\mathcal J
\bigl(p_{\mathrm{txt}} (\mathcal C_i, \widetilde y_{1,i,j}) \bigr)
\in\{0,1\}^{K_i},
\label{eq:online-text-judge}
\end{equation}
where \(b_{i,j,k}=1\) indicates that the CoT satisfies criterion \(c_{i,k}\).
With \(\mathbf w_i=(w_{i,1},\ldots,w_{i,K_i})\), the rubric reward is:
\begin{equation}
R_{i,j}^{\mathrm{RUB}}
=\dfrac{\mathbf w_i^\top\mathbf b_{i,j}}{\lVert\mathbf w_i\rVert_1}
\in[0,1].
\label{eq:rubric-scores}
\end{equation}
The final rollout reward is
\begin{equation}
R_{i,j}=R_{i,j}^{\mathrm{MCQ}}R_{i,j}^{\mathrm{RUB}}.
\label{eq:rubric-reward}
\end{equation}

Thus, trajectory correctness determines whether a rollout receives process supervision, while neither the GT trajectory nor the candidate set serves as input to the Turn-1 causal reasoning process.
\textbf{Crucially, \(\mathbf P_i^{\mathrm{gt}}\) is never directly provided to $\mathcal G $, \(\pi_\theta\)  or \(\mathcal J\).}
The judge evaluates the reasoning process solely according to the predefined rubric criteria, \textbf{without direct access to the visual input, candidate options, or GT trajectory}.
Compared with directly providing a VLM-based judge with the full visual context, this text-only grading scheme is substantially more efficient.
Moreover, it allows the judge to focus on assessing the quality of the textual reasoning, without its attention being diluted by a large number of visual tokens \cite{zhou2025libero,fei2025libero}.

\section{Experiments}


\begin{table*}[t]
\centering
\small
\setlength{\tabcolsep}{1mm}
\resizebox{\textwidth}{!}{%
\begin{tabular}{@{}lcccccccc@{}}
\specialrule{1pt}{0pt}{\belowrulesep}
\multirow{2}{*}{\textbf{Method}}
& \multicolumn{3}{c}{\textbf{AD-Specific Reasoning}}
& \multicolumn{5}{c}{\textbf{General Visual Capability(\%)}} \\
\cmidrule(lr){2-4}\cmidrule(lr){5-9}
& \textbf{ACC(\%)} $\uparrow$
& \textbf{CFS} $\uparrow$
& \textbf{HLD} $\uparrow$
& \textbf{Basic} $\uparrow$
& \textbf{Embodied} $\uparrow$
& \textbf{3D/MV} $\uparrow$
& \textbf{RefSpatial} $\uparrow$
& \textbf{Avg.} $\uparrow$ \\
\midrule
\textbf{Qwen3-VL-8B-Instruct} & 28.1 & -- & -- & \multirow{2}{*}{81.60} & \multirow{2}{*}{56.63} & \multirow{2}{*}{42.48} & \multirow{2}{*}{38.52} & \multirow{2}{*}{54.81} \\
\cmidrule(lr){1-4}
+ DEFT & \textbf{56.6} & 0.431 & 0.425 & & & & & \\
\midrule
+ JEFT $+$ RLVR ($R^{\mathrm{MCQ}}$) & 61.1 & 0.428 & 0.427 & \underline{81.39} & 56.67 & 41.99 & 39.56 & 54.90 \\
+ DEFT $+$ RLVR ($R^{\mathrm{MCQ}}$) & \underline{76.4} & 0.442 & 0.462 & 81.36 & \underline{57.15} & \underline{42.80} & \underline{44.26} & \textbf{56.39} \\
+ DEFT $+$ RLVR ($R^{\mathrm{MCQ}}R^{\mathrm{GEN}}$) & 75.2 & \underline{0.580} & \underline{0.487} & \textbf{81.59} & 56.31 & \textbf{42.81} & \textbf{44.79} & \underline{56.37} \\
+ DEFT-RLVR & \textbf{77.9} & \textbf{0.658} & \textbf{0.501} & 81.33 & \textbf{57.41} & 42.27 & 43.35 & 56.09 \\
\midrule
+ JEFT Distillation & 64.0 & 0.620 & 0.480 & 74.20 & 52.80 & 39.60 & 32.60 & 49.80 \\
+ DEFT Distillation (Plan Only) & 68.2 & \underline{0.925} & 0.560 & \textbf{78.00} & 53.49 & \underline{40.20} & 33.46 & \underline{51.29} \\
+ DEFT Distillation (Full Interaction) & \underline{82.4} & 0.909 & \underline{0.591} & 74.94 & \textbf{54.04} & \textbf{40.38} & \underline{34.56} & 50.98 \\
+ DEFT Distillation (Mixed Targets) & \textbf{84.1} & \textbf{0.934} & \textbf{0.627} & \underline{76.73} & \underline{53.51} & \underline{40.20} & \textbf{36.76} & \textbf{51.80} \\
\specialrule{1pt}{\aboverulesep}{\belowrulesep}
\textbf{Qwen3.5-4B} & 34.0 & -- & -- & \multirow{2}{*}{80.31} & \multirow{2}{*}{53.00} & \multirow{2}{*}{40.73} & \multirow{2}{*}{36.19} & \multirow{2}{*}{52.56} \\
\cmidrule(lr){1-4}
+ DEFT & \textbf{65.6} & 0.738 & 0.481 & & & & & \\
\midrule
+ JEFT $+$ RLVR ($R^{\mathrm{MCQ}}$) & 72.3 & 0.740 & 0.511 & 80.23 & 53.34 & 40.16 & 31.32 & 51.26 \\
+ DEFT $+$ RLVR ($R^{\mathrm{MCQ}}$) & 79.0 & 0.805 & 0.529 & \underline{80.76} & 52.75 & \textbf{41.83} & \textbf{38.72} & \textbf{53.52} \\
+ DEFT $+$ RLVR ($R^{\mathrm{MCQ}}R^{\mathrm{GEN}}$) & \underline{79.4} & \underline{0.819} & \underline{0.540} & 80.41 & \underline{53.37} & 40.67 & \underline{38.62} & \underline{53.27} \\
+ DEFT-RLVR  & \textbf{82.2} & \textbf{0.822} & \textbf{0.582} & \textbf{80.88} & \textbf{55.20} & \underline{40.94} & 35.93 & 53.24 \\
\specialrule{1pt}{\aboverulesep}{0pt}
\end{tabular}
}
\caption{Main results on AD reasoning and general visual capability.
Base models use JEFT for AD evaluation.
CFS and HLD are scored by Qwen3.5-397B-A17B, with strong agreement with human
annotations demonstrated in Appendix~\ref{app:human-judge-validation}.}
\label{tab:main-results}
\end{table*}

\subsection{Experimental Setup}
\label{sec:exp-setup}

\paragraph{Data and benchmark.}
We divide scenes from Waymo Open E2E \cite{xu2026wod} and an internal driving corpus into
Train, Dev, and \textbf{AD-MCQ-500} Test splits, containing
$5{,}000$, $100$, and $500$ scenes, respectively, with \(K{=}8192\) 
and \(N{=}489{,}042\).
Each visual input \(V_i\) contains four frames sampled at \(2\,\mathrm{Hz}\)
from three cameras: front-left, front, and front-right.
Train follows the natural scene distribution, whereas Dev and Test focus on
causally demanding scenes with structured hard distractors.
Dev is curated as the harder of the two evaluation splits.
Details are shown in Appendix~\ref{app:ad-mcq-construction}.

\paragraph{Evaluation.}
For AD-specific evaluation, we report \textbf{ACC} (\textbf{AD-MCQ-500} accuracy)   and two
complementary CoT metrics:
\textbf{CFS} (Normalized Causal-Faithfulness Score) and
\textbf{HLD} (High-Level-Decision Consistency).

To assess general-capability retention, we use $12$ vision-language benchmarks
covering four capability groups \cite{duan2024vlmevalkit,yu2026vulcabench}: basic visual perception
(\textbf{Basic Visual}) \cite{tong2024cambrian,yang2024depth,sun2026probing}, embodied spatial reasoning
(\textbf{Embodied Spatial}) \cite{du2024embspatial, song2024robospatial,team2025gemini,ma2026thinkingblueprintsassistingvisionlanguage}, 3D and multi-view reasoning
(\textbf{3D/Multi-View}) \cite{ma20253dsrbench,yang2025mmsi,li2025viewspatial,wang2026verybigvideo}, and referring-expression-based spatial grounding
(\textbf{RefSpatial}) \cite{zhou2026roborefer}.
We additionally evaluate cross-domain AD transfer on an external
$500$-scene nuScenes set.
Complete evaluation settings are provided in Appendix~\ref{app:evaluation-settings}.

\paragraph{Models.}
We use Qwen3-VL-8B-Instruct \cite{bai2025qwen3} and Qwen3.5-4B \cite{qwen3.5} as the base models.
Qwen3.5-397B-A17B provides supervision targets for distillation,
while Qwen3.6-35B-A3B generates instance-specific rubrics offline and serves
as the reasoning-process judge.

\paragraph{Candidate-Exposure Settings.}
\textbf{DEFT} (Deferred Exposure of Future
Trajectories) first elicits a pre-exposure plan and reveals the candidate
trajectories only in the subsequent selection turn.
Conversely, \textbf{JEFT} (Joint Exposure of Future Trajectories) is the matched
ablation that presents the same scene context and candidate
trajectories jointly with the same prompts.
For a fair comparison, both settings use $T{=}1.0$, top-$p{=}0.95$, and the
same total token budget of $24{,}576$ tokens, with DEFT capped at $12{,}000$
tokens per turn.

\paragraph{RLVR Variants.}

\textbf{JEFT $+$ RLVR ($R^{\mathrm{MCQ}}$)} and
\textbf{DEFT $+$ RLVR ($R^{\mathrm{MCQ}}$)} use the same exact-choice reward
and differ in candidate-exposure order.
\textbf{DEFT $+$ RLVR
($R^{\mathrm{MCQ}}R^{\mathrm{GEN}}$)} uses a shared rubric whose score
$R^{\mathrm{GEN}}$ is assigned online by a scene-conditioned VLM
grader.
\textbf{DEFT-RLVR} uses the instance-specific reward
$R^{\mathrm{MCQ}}R^{\mathrm{RUB}}$ defined in Eq.~\ref{eq:rubric-reward}.

\paragraph{Distillation Variants.}
We prompt Qwen3.5-397B-A17B under the corresponding exposure setting and use
its responses as supervised fine-tuning targets for the student.
\textbf{JEFT Distillation} imitates the single-turn reasoning-and-selection
response generated with candidates exposed from the outset.
\textbf{DEFT Distillation (Plan Only)} imitates only the Turn-1 plan generated
before candidate exposure.
\textbf{DEFT Distillation (Full Interaction)} supervises the complete two-turn
plan-then-match interaction.
\textbf{DEFT Distillation (Mixed Targets)} uses an equal mixture of plan-only
and full-interaction targets.
Detailed implementations of the RLVR and distillation variants are provided in
Appendix~\ref{app:baselines-controlled-variants}.

\subsection{Candidate-Grounded Training Improves Generalizable AD Reasoning}
\label{sec:exp-ad-reasoning}

\paragraph{DEFT mitigates candidate anchoring bias across
inference and training.}
As illustrated in Table~\ref{tab:main-results},
Compared with training-free \textbf{JEFT}, \textbf{DEFT} raises ACC from $28.1\%$ to $56.6\%$ on Qwen3-VL-8B
and from $34.0\%$ to $65.6\%$ on Qwen3.5-4B.
Under the same correctness-only reward,
\textbf{DEFT $+$ RLVR ($R^{\mathrm{MCQ}}$)} exceeds its matched
\textbf{JEFT $+$ RLVR ($R^{\mathrm{MCQ}}$)} ablation by $15.3$ and $6.7$ percentage points on Qwen3-VL-8B and Qwen3.5-4B, respectively.
Across both backbones, DEFT-RLVR jointly improves MCQ accuracy,
CFS, and HLD over training-free DEFT, demonstrating gains in
reasoning quality and decision consistency rather than final-choice accuracy
alone.
Under equal-data distillation, \textbf{DEFT Distillation (Full Interaction)}
outperforms \textbf{JEFT Distillation}, raising MCQ accuracy from $64.0\%$ to
$82.4\%$, while improving CFS by $0.289$ and HLD by $0.111$.

These consistent performance gaps can be attributed to the same
information-order mechanism: 
when future-trajectory candidates are visible
during decision formation, the policy may organize the
reasoning around a favored answer, creating a shortcut consistent with the
anchoring bias shown in Section~\ref{sec:motivation}.
DEFT prevents this shortcut by requiring the model to derive its decision from
scene evidence before grounding it in a concrete trajectory.

\paragraph{Fine-grained trajectory grounding strengthens generalizable AD reasoning.}
Compared with the Plan Only variant,
\textbf{DEFT Distillation (Mixed Targets)} incorporates full two-turn targets
and raises accuracy from $68.2\%$ to $84.1\%$, CFS from $0.925$ to
$0.934$, and HLD from $0.560$ to $0.627$.
These simultaneous gains show that 
Turn-2 trajectory grounding is
more than a mechanism for providing RL with an exact, verifiable reward:
the required fine-grained discrimination among trajectory candidates also
improves reasoning quality and trajectory-decision accuracy.

\paragraph{Candidate-grounded training delivers AD reasoning gains that generalize
to an OOD driving domain.}

On the out-of-distribution (OOD) nuScenes domain
\cite{caesar2020nuscenes}, both DEFT training variants still significantly outperform training-free DEFT on
all three AD metrics.
As shown in Table~\ref{tab:nuscenes-generalization}, DEFT-RLVR raises candidate accuracy from $39.6\%$ to $49.5\%$,
while improving CFS by $0.114$ and HLD by $0.073$.

Together, these gains indicate that our training paradigm helps the policy learn
transferable scene-to-decision reasoning and subsequent explicit-trajectory
grounding, rather than rely on source-specific visual cues.
\begin{table}[h]
\centering
\small
\setlength{\tabcolsep}{1mm}
\begin{tabular}{@{}lccc@{}}
\toprule
\textbf{Method} & \textbf{ACC} $\uparrow$ & \textbf{CFS} $\uparrow$ & \textbf{HLD} $\uparrow$ \\
\midrule
\textbf{DEFT (Training-Free)} & 39.6 & 0.522 & 0.286 \\
\textbf{DEFT Distillation (Mixed Targets)} & 55.8 & 0.655 & 0.370 \\
\textbf{DEFT-RLVR} & 49.5 & 0.636 & 0.359 \\
\bottomrule
\end{tabular}
\caption{Cross-domain results on 500 nuScenes scenes using
Qwen3-VL-8B-Instruct. Both candidate-grounded training variants outperform
training-free DEFT across accuracy (ACC), causal-faithfulness score (CFS), and
high-level decision consistency (HLD), demonstrating that the learned
scene-to-decision reasoning transfers beyond the training domain. Mixed-target
distillation achieves the strongest overall results, while DEFT-RLVR also
delivers consistent gains using verifiable reward supervision.}
\label{tab:nuscenes-generalization}
\end{table}


\begin{figure}[h]
\centering
\includegraphics[width=0.6\columnwidth]{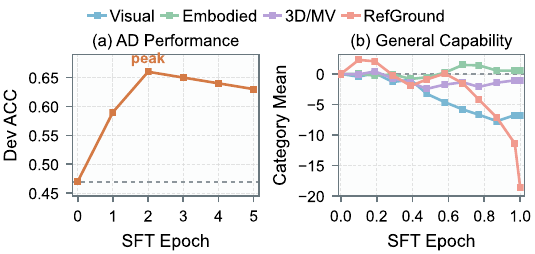}
\caption{Cold-start SFT trades general capability for AD specialization.
(a) Hard-100 Dev accuracy.
(b) First-epoch performance changes across four general-capability groups
relative to the Base VLM.}
\label{fig:coldstart-main}
\end{figure}

\begin{figure*}[t]
\centering
\includegraphics[width=1.0\textwidth]{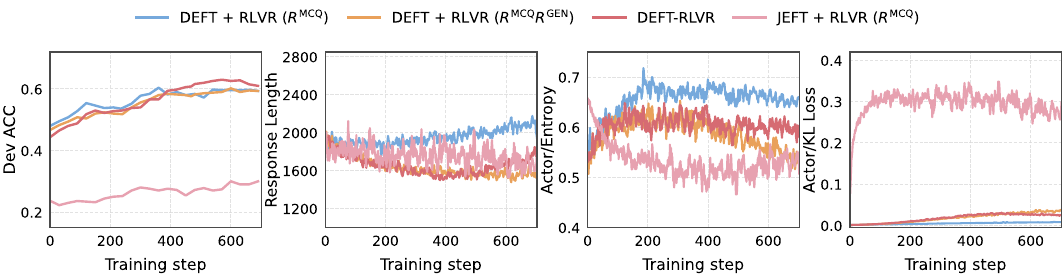}
\caption{Training dynamics of the Qwen3-VL-8B-Instruct RLVR variants. From
left to right, the panels report development-set accuracy, response length,
actor entropy, and actor KL loss. Candidate-visible JEFT exhibits by far the
largest policy drift while remaining the least accurate. Deferred exposure
substantially improves accuracy, and rubric-supervised DEFT-RLVR attains the
strongest late-stage performance while keeping responses shorter and entropy
lower than correctness-only DEFT, indicating more controlled and productive
exploration.}
\label{fig:rl-dynamics-four}
\end{figure*}

\subsection{RLVR Improves AD Reasoning without Sacrificing General Visual Capability}
\label{sec:exp-specialization-retention}

\paragraph{Distillation improves AD specialization at the cost of general capability.}
As shown in Table~\ref{tab:main-results}, JEFT Distillation reduces the average of general visual capability from $54.81\%$ to $49.80\%$, while the
equal-data DEFT variants retain $50.98\%$--$51.80\%$.
This is because token-level SFT supervision pushes the student toward an
AD-specific response distribution generated by an external teacher rather than
selectively reinforcing correct behavior.
Although DEFT-based distillation mitigates this policy shift relative to
shortcut-prone JEFT distillation, dense teacher imitation still trades general
capability for AD performance.

\paragraph{Cold-start SFT causes an early decline in general visual capability.}
To assess whether RLVR should start from an AD-specialized policy, we first
apply SFT to teacher-generated responses as a cold-start stage, with experimental details shown in
 Appendix~\ref{app:cold-start-sft}.
As shown in Figure~\ref{fig:coldstart-main}, 
although cold-start SFT improves Dev accuracy, 
all four general-capability groups decline from the first epoch. 
Initializing RL from this policy would additionally anchor KL regularization to an already shifted policy.
Thus, we start DEFT-RLVR from the unmodified Base VLM.
\paragraph{DEFT-RLVR improves AD reasoning while preserving general visual capability.}
As illustrated in Table \ref{tab:main-results},  DEFT-RLVR raises the average of general visual capability from $54.81\%$ to $56.09\%$ on Qwen3-VL-8B and
from $52.56\%$ to $53.24\%$ on Qwen3.5-4B.
Unlike SFT, RL-based variants learn from responses sampled from the current or a recent policy. The resulting policy gradients merely increase or decrease the probability of each sampled token conditioned on its corresponding context \cite{zhu2026surprising}, thereby constituting a more fine-grained form of policy optimization than SFT \cite{fu2025srft}.
Meanwhile, the causal reasoning process partially  exercises visual-spatial reasoning shared with the
general benchmarks, which may explain the modest gains of general visual capability.
The evaluation results of all 12 benchmarks are provided in Appendix \ref{app:general-visual-results}.

\subsection{Ablation of the RLVR Design}
\label{sec:exp-rl-ablation}

\paragraph{JEFT $+$ RLVR ($R^{\mathrm{MCQ}}$) vs.\ DEFT $+$ RLVR
($R^{\mathrm{MCQ}}$): deferred exposure avoids shortcut-driven optimization.}
Figure~\ref{fig:rl-dynamics-four} shows that the JEFT variant exhibits substantially larger policy drift while remaining less accurate than DEFT variants.
As shown in Table~\ref{tab:main-results}, DEFT improves accuracy from $61.1\%$ to $76.4\%$ on Qwen3-VL-8B and from $72.3\%$ to $79.0\%$ on Qwen3.5-4B, while also achieving higher CFS and HLD.
These results suggest that deferred exposure effectively reduces candidate-visible shortcuts and promotes more effective scene-derived reasoning.

\paragraph{DEFT $+$ RLVR ($R^{\mathrm{MCQ}}$) vs.\ DEFT-RLVR:
rubric supervision prunes unproductive exploration.}
As shown in Figure~\ref{fig:rl-dynamics-four}, DEFT $+$ RLVR
($R^{\mathrm{MCQ}}$) produces the longer and higher-entropy responses than 
rubric-supervised variants.
Meanwhile, Table \ref{tab:main-results} shows that the introduction of  rubric supervision consistently  enhances the reasoning and HLD quality  on both backbones.
This improvement can be attributed to the rubric’s fine-grained supervision, which effectively steers the reasoning process toward greater faithfulness while suppressing unproductive exploration.

\paragraph{DEFT-RLVR vs. DEFT $+$ RLVR ($R^{\mathrm{MCQ}}R^{\mathrm{GEN}}$):
stronger reasoning capacity with marginal time cost.}
As shown in Table~\ref{tab:reward-cost}, compared with DEFT $+$ RLVR ($R^{\mathrm{MCQ}}$),
DEFT-RLVR substantially improves reasoning quality while
introducing marginal training cost, increasing total step
time by only $0.5\%$, from $424.5$ to $426.5$ seconds.
DEFT-RLVR likewise achieves higher reasoning quality than the online-rubric variant, DEFT $+$ RLVR ($R^{\mathrm{MCQ}}R^{\mathrm{GEN}}$), despite incurring substantially lower training costs.
Specifically, by constructing instance-specific criteria offline and retaining only
text-based grading online, DEFT-RLVR reduces total step time from $724.4$
to $426.5$ seconds ($41.1\%$ faster).

\begin{table}[t]
\centering
\small
\setlength{\tabcolsep}{5pt}
\begin{tabular}{@{}lccc@{}}
\hline
Method & Step $\downarrow$ & Rollout $\downarrow$ & Scoring $\downarrow$ \\
\hline
\textbf{DEFT $+$ RLVR ($R^{\mathrm{MCQ}}$)} & \textbf{424.5} & \textbf{171.1} & \textbf{0.03} \\
\textbf{DEFT $+$ RLVR ($R^{\mathrm{MCQ}}R^{\mathrm{GEN}}$)} & 724.4 & 489.8 & 156.8 \\
\textbf{DEFT-RLVR} & \underline{426.5} & \underline{174.6} & \underline{4.12} \\
\hline
\end{tabular}
\caption{Per-step runtime of Qwen3-VL-8B-Instruct DEFT RLVR variants in seconds.
DEFT-RLVR adds only $0.5\%$ overhead over correctness-only training while being
$41.1\%$ faster than the online-rubric variant, demonstrating its superior
efficiency for rubric-supervised optimization.}
\label{tab:reward-cost}
\end{table}


\begin{figure}[t]
\centering
\includegraphics[width=0.45\columnwidth]{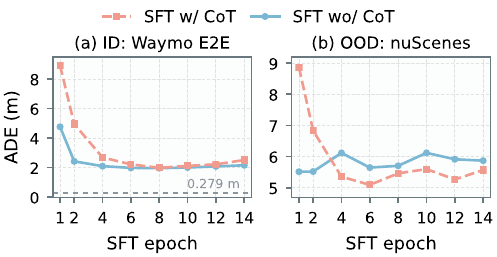}
\includegraphics[width=0.45\columnwidth]{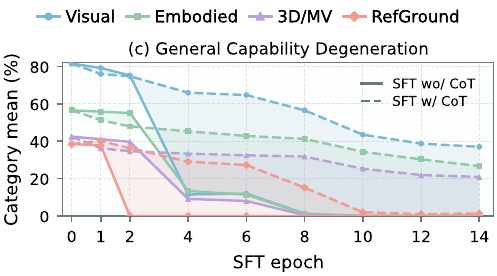}
\caption{Direct trajectory-token generation introduces coupled prediction and
capability-retention bottlenecks. (a,b) Despite increasing fit under SFT, both
in-distribution and out-of-distribution trajectory-prediction ADEs remain well
above the codebook reconstruction floor of $0.279\,$m, indicating that most of
the error arises from token inference rather than trajectory quantization.
(c) Direct token supervision also substantially degrades general visual
capability, and incorporating CoT does not prevent this degradation. Together,
these results motivate externalizing planning as selection over explicit
trajectory candidates rather than internalizing a large trajectory-token
vocabulary.}
\label{fig:traj-overfit}
\end{figure}

\subsection{Why Formulate AD Planning as a Candidate-Grounded MCQ?}
\label{sec:exp-why-mcq}

\paragraph{Candidate grounding avoids the dual bottleneck of trajectory error and
general-capability degradation.}
Using a shared codebook, we fine-tune Qwen3-VL-8B-Instruct via SFT to predict
trajectory tokens with or without trajectory-conditioned CoT; full details are
provided in Appendix~\ref{app:direct-token-sft}.
Figure~\ref{fig:traj-overfit}(a,b) shows that, even after SFT begins to overfit, both in- and out-of-distribution prediction ADEs remain substantially above the codebook's $0.279\,$m reconstruction ADE, 
identifying token generation as the primary error source.
Figure~\ref{fig:traj-overfit}(c) further indicates that direct trajectory generation
severely degrades general visual capability even with CoT.
Both failures arise because training forces the VLM to internalize a large
trajectory-token inventory within its original vocabulary, substantially
perturbing the pretrained token distribution.
AD-MCQ instead uses the codebook only to retrieve waypoint candidates and
externalizes generation as selection among scene-conditioned explicit
trajectories, thereby avoiding both bottlenecks.

\begin{figure}[!ht]
\centering
\includegraphics[width=0.6\columnwidth]{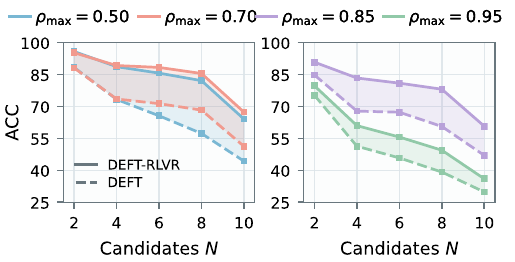}
\caption{Robustness to candidate-set construction. Holding scenes and oracle
trajectories fixed, we resample distractors across five candidate-set sizes and
four hard-negative similarity bounds. DEFT-RLVR consistently outperforms
training-free DEFT in all $20$ configurations, indicating that its gains
transfer across candidate constructions rather than depending on a fixed
distractor geometry.}
\label{fig:candidate-set-robustness}
\end{figure}

\paragraph{The gains of DEFT-RLVR generalize across diverse MCQ option constructions.}
Holding scenes and oracle trajectories fixed, we randomly resample distractors
across five candidate counts and four hard-negative similarity bounds.
Figure~\ref{fig:candidate-set-robustness} shows that DEFT-RLVR
outperforms training-free DEFT in all $20$ settings.
Thus, the capability learned by DEFT-RLVR under a fixed MCQ configuration
transfers to new candidate-set constructions rather than relying on a
particular distractor geometry.

Meanwhile, highly similar future trajectories and larger candidate sets
remain the most challenging regimes for fine-grained candidate grounding.
Through controlled option construction, AD-MCQ thus provides a simple yet
difficulty-controllable experimental paradigm for future research.
Appendix~\ref{app:candidate-set-ablation} provides further experimental
details.

\section{Conclusion}

We identify anchoring bias in AD VLMs, propose AD-MCQ and leverage DEFT-RLVR for training. 
This framework improves generalizable AD reasoning while preserving and even enhancing the model's general visual capabilities. 
Since AD-MCQ relies solely on the VLM and allows difficulty to be controlled through option construction, 
it provides a highly deployable and scalable foundation for future research.

\clearpage
\bibliographystyle{assets/plainnat}
\bibliography{references}

\clearpage
\appendix
\renewcommand{\thesection}{\Alph{section}}
\section*{Appendix}
\tableofcontents
\clearpage
\section{Related Work}
\label{app:related-work}

\subsection{Vision--Language Reasoning for Autonomous Driving}

Language-conditioned driving models have evolved from using language as an
auxiliary source of supervision to placing a VLM directly in the perception,
reasoning, and planning loop. Early systems formulate driving as graph-based
visual question answering, language-conditioned behavior prediction, or
interpretable trajectory generation
\cite{sima2024drivelm,mao2023gpt,wang2023drivemlm,xu2024drivegpt4}.
Subsequent work expands this direction through multi-view scene reasoning,
knowledge augmentation, behavioral planning states, and unified
vision--language--action architectures
\cite{tian2024drivevlm,zhang2024wisead,jiang2024senna,
hwang2024emma,zhou2025opendrivevla}.
Driving-oriented question-answering and reasoning benchmarks complement these
models by measuring scene understanding, spatial reasoning, and interpretable
decision making
\cite{marcu2024lingoqa,park2025nuplanqa,qian2024nuscenes,
tian2025nuscenes,nie2024reason2drive,li2024womd}.

More recent methods explicitly supervise chain-of-thought reasoning or combine
reasoning traces with action learning
\cite{wang2024drivecot,zhao2025cot,zhou2026autovla,ishaq2025drivelmm,
wei2025ad,gu2026accelerating}.
These approaches establish the value of explicit intermediate reasoning for
driving. Our work addresses a distinct question concerning the direction of
that supervision: when a rationale is generated with access to the logged
future trajectory, the trajectory can become a premise from which the teacher
works backward. We instead require a scene-grounded decision before exposing
future-trajectory candidates, retaining trajectory-level supervision while
preventing the target future from anchoring the initial reasoning process.

\subsection{Trajectory Representations and End-to-End Planning}

End-to-end driving has been studied through sensor-fusion policies,
planning-oriented representations, sparse scene abstractions, and integrated
prediction--planning architectures
\cite{chitta2022transfuser,hu2023planning,jiang2023vad,sun2024sparsedrive,
huang2023differentiable,huang2023gameformer}.
Generative planners further model multimodal futures with autoregressive,
diffusion, or flow-based objectives
\cite{zheng2024genad,liao2024diffusiondrive,huang2024gen,tan2026flow}.
In parallel, discretized action and trajectory representations make continuous
behavior compatible with token-based sequence models
\cite{philion2023trajeglish,wu2024smart,zhang2024closed,pertsch2025fast,
li2025discrete}.
Such representations reduce the mismatch between language-model decoding and
continuous control, but direct full-vocabulary trajectory generation still
requires the VLM to synthesize precise geometry and can encourage
task-specific memorization.

AD-MCQ uses trajectory prototypes differently. The prototypes define a
scene-specific set of decoded, explicit candidates rather than a global
action vocabulary that the VLM must generate. This formulation preserves
differences in lateral geometry, braking time, and speed profile while turning
trajectory-level planning into exact candidate selection. It is therefore
closer to a verification interface than to a replacement for a downstream
continuous planner. Our candidate-construction and held-out representation
analyses further separate codebook sufficiency from the difficulty of
full-vocabulary trajectory-token prediction.

\subsection{Verifiable Post-Training and Faithful Reasoning}

Supervised instruction tuning, preference optimization, and reinforcement
learning provide complementary mechanisms for adapting foundation models
\cite{ouyang2022training,rafailov2023direct,schulman2017proximal,wu2026star,wu2024separate,ma2025styletailorpersonalizedfashionstyling,Ma_Chen_Zhang_Wu_Ding_2025,huang2025adaptive}.
For reasoning models, reinforcement learning with automatically checkable
outcomes can elicit capabilities without requiring imitation of every
intermediate step
\cite{shao2024deepseekmath,guo2025deepseek,yang2025qwen3,yue2026promoting,liu2026automated,huang2026does,yang2026batched,zhang2026heterogeneous}.
Recent work extends this principle beyond exact symbolic answers through
multidimensional rubrics and rubric-derived rewards
\cite{hashemi2024llm,rao2026autorubric,gunjal2025rubrics,rezaei2025online,Huang2026RealTimeAR},
while studies of multiple-choice RLVR show that distractor construction and
answer format materially shape the learned behavior
\cite{guo2026rethinking,chandak2025answer}.
DEFT-RLVR combines exact candidate correctness with instance-specific rubric
rewards, but gates process rewards on outcome correctness and grades the
candidate-blind reasoning trace without visual or answer-related information.

This design is also motivated by evidence that a chain of thought need not be a
faithful account of the evidence that produced an answer. Preemptively
revealing an answer can distort subsequent reasoning, and both language and
vision--language models may rationalize cues or hallucinate support for an
already favored conclusion
\cite{xu-etal-2024-preemptive,arcuschin2025chain,liu2026toolanchor,
balasubramanian-etal-2025-closer,chen2025reasoning,xu2026more}.
Our controlled annotation study instantiates this issue in trajectory-level
driving decisions, and our deferred-exposure formulation converts the future
trajectory from a pre-reasoning cue into a post-decision verification target.

\clearpage

\newcommand{\DEFTRLVRTrainingAlgorithm}{%

Algorithm~\ref{alg:deft-rlvr} integrates the four RLVR configurations
into a shared on-policy training loop. Color-coded branches isolate their
candidate-exposure interfaces and reward computations, while all black steps
use the same rollout grouping and GRPO update.

\refstepcounter{deftalgorithm}
\label{alg:deft-rlvr}
\begin{tcolorbox}[
  colback=white,
  colframe=black!55,
  boxrule=0.6pt,
  arc=0pt,
  left=6pt,
  right=6pt,
  top=6pt,
  bottom=6pt
]
\noindent\textbf{Algorithm \thedeftalgorithm: Unified On-Policy Training for the RLVR Variants}
\par\smallskip\hrule\smallskip
\small
\begin{tabular}{@{}r@{\hspace{0.7em}}p{\dimexpr\linewidth-2.5em\relax}@{}}
\multicolumn{2}{@{}p{\linewidth}@{}}{\textbf{Input:}
training instances
$\mathcal D=\{(X_i,\mathcal A_i,a_i^\star)\}$;
variant $v$;
policy $\pi_\theta$;
rubric generator $\mathcal G$;
text grader $\mathcal J$;
prompts $p_1,p_2,p_{\mathrm{rub}},p_{\mathrm{txt}}$;
rollouts per instance $G$.}\\
\multicolumn{2}{@{}p{\linewidth}@{}}{\textbf{Variants:}
\textcolor{jeftpink}{\textbf{JEFT $+$ RLVR ($R^{\mathrm{MCQ}}$)}};
\textcolor{deftmcqblue}{\textbf{DEFT $+$ RLVR ($R^{\mathrm{MCQ}}$)}};
\textcolor{deftgenorange}{\textbf{DEFT $+$ RLVR
($R^{\mathrm{MCQ}}R^{\mathrm{GEN}}$)}};
\textcolor{deftrlvrred}{\textbf{DEFT-RLVR}}.}\\
\multicolumn{2}{@{}p{\linewidth}@{}}{\textbf{Output:} optimized policy $\pi_\theta$.}\\[2pt]
1 & \textcolor{deftrlvrred}{Generate and cache
    $\mathcal C_i\leftarrow\mathcal G(p_{\mathrm{rub}}(X_i))$
    for every $i\in\mathcal D$.}\\
2 & \textbf{for each on-policy RL iteration} $t=1,2,\ldots$ \textbf{do}\\
3 & \hspace*{1em}Sample a fresh minibatch $\mathcal B_t\subset\mathcal D$ and
    set the rollout policy $\pi_{\mathrm{old}}\leftarrow\pi_{\theta_t}$.\\
4 & \hspace*{1em}\textbf{for each} $i\in\mathcal B_t$ and $j\in\{1,\ldots,G\}$ \textbf{do}\\
5 & \deftvariantbox{jeftpink}{%
    \textcolor{jeftpink}{\textbf{JEFT $+$ RLVR
    ($R^{\mathrm{MCQ}}$):}} Reveal $\mathcal A_i$ at the outset and sample one
    joint reasoning-and-selection response $y_{i,j}$ from
    $\pi_{\mathrm{old}}$.}\\[2pt]
6 & \deftvariantbox{deftsharedgray}{%
    \textbf{All DEFT variants}
    [\textcolor{deftmcqblue}{\textbf{DEFT $+$ RLVR ($R^{\mathrm{MCQ}}$)}};
    \textcolor{deftgenorange}{\textbf{DEFT $+$ RLVR
    ($R^{\mathrm{MCQ}}R^{\mathrm{GEN}}$)}};
    \textcolor{deftrlvrred}{\textbf{DEFT-RLVR}}]:
    sample the candidate-free $y_{1,i,j}$; then reveal $\mathcal A_i$ and
    sample the matching response $y_{2,i,j}$.}\\[2pt]
7 & \hspace*{2em}Parse $\widehat a_{i,j}$ from the final response and set
    $R_{i,j}^{\mathrm{MCQ}}\leftarrow
    \mathbb I[\widehat a_{i,j}=a_i^\star]$.\\
8 & \deftvariantbox{jeftpink}{%
    \textcolor{jeftpink}{\textbf{JEFT $+$ RLVR
    ($R^{\mathrm{MCQ}}$):}}
    $R_{i,j}\leftarrow R_{i,j}^{\mathrm{MCQ}}$.}\\[2pt]
9 & \deftvariantbox{deftmcqblue}{%
    \textcolor{deftmcqblue}{\textbf{DEFT $+$ RLVR
    ($R^{\mathrm{MCQ}}$):}}
    $R_{i,j}\leftarrow R_{i,j}^{\mathrm{MCQ}}$; assign it to both turns.}\\[2pt]
10 & \deftvariantbox{deftgenorange}{%
     \textcolor{deftgenorange}{\textbf{DEFT $+$ RLVR
     ($R^{\mathrm{MCQ}}R^{\mathrm{GEN}}$):}}
     If $R_{i,j}^{\mathrm{MCQ}}=1$, grade $\widetilde y_{1,i,j}$ together
     with the scene frames using the shared image-conditioned rubric to obtain
     $R_{i,j}^{\mathrm{GEN}}$; otherwise set $R_{i,j}^{\mathrm{GEN}}=0$.
     Then set
     $R_{i,j}\leftarrow
     R_{i,j}^{\mathrm{MCQ}}R_{i,j}^{\mathrm{GEN}}$.}\\[2pt]
11 & \deftvariantbox{deftrlvrred}{%
     \textcolor{deftrlvrred}{\textbf{DEFT-RLVR:}}
     If $R_{i,j}^{\mathrm{MCQ}}=1$, grade $\widetilde y_{1,i,j}$ against $\mathcal C_i$
     with the text-only grader and compute
     $R_{i,j}^{\mathrm{RUB}}=
     \mathbf w_i^\top\mathbf b_{i,j}/\lVert\mathbf w_i\rVert_1$;
     otherwise set $R_{i,j}^{\mathrm{RUB}}=0$.
     If grading fails after retries, set
     $R_{i,j}^{\mathrm{RUB}}\leftarrow1$.
     Then set
     $R_{i,j}\leftarrow
     R_{i,j}^{\mathrm{MCQ}}R_{i,j}^{\mathrm{RUB}}$.}\\[2pt]
12 & \hspace*{2em}Serialize the generated response or complete two-turn
     interaction as $s_{i,j}$.\\
13 & \hspace*{1em}\textbf{end for}\\
14 & \hspace*{1em}Normalize $\{R_{i,j}\}_{j=1}^{G}$ within each rollout
     group to obtain the clipped advantages $\{\widehat A_{i,j}\}_{j=1}^{G}$.\\
15 & \hspace*{1em}Compute generated-token importance ratios between
     $\pi_{\theta}$ and $\pi_{\mathrm{old}}$ over $\{s_{i,j}\}$, with prompt
     tokens masked and $\widehat A_{i,j}$ shared by all generated turns.\\
16 & \hspace*{1em}Take exactly one GRPO update
     $\theta_t\rightarrow\theta_{t+1}$, applying KL regularization separately
     from the reward.\\
17 & \hspace*{1em}Discard the rollout batch; do not reuse it after the policy
     update.\\
18 & \textbf{end for} when the training budget is exhausted.\\
\end{tabular}
\end{tcolorbox}
}

\newcommand{\DistillationTrainingAlgorithm}{%

Algorithm~\ref{alg:distillation} summarizes the controlled teacher-target
construction and student fine-tuning pipeline. All variants use the same
teacher-labeled scene budget and student initialization; they differ only in
when candidate trajectories are exposed to the teacher and which generated
turns are retained as supervised targets.

\refstepcounter{deftalgorithm}
\label{alg:distillation}
\begin{tcolorbox}[
  colback=white,
  colframe=black!55,
  boxrule=0.6pt,
  arc=0pt,
  left=6pt,
  right=6pt,
  top=6pt,
  bottom=6pt
]
\noindent\textbf{Algorithm \thedeftalgorithm: Unified Target Construction and Training for the Distillation Variants}
\par\smallskip\hrule\smallskip
\small
\begin{tabular}{@{}r@{\hspace{0.7em}}p{\dimexpr\linewidth-2.5em\relax}@{}}
\multicolumn{2}{@{}p{\linewidth}@{}}{\textbf{Input:}
training instances
\(\mathcal D=\{(X_i,\mathcal A_i,a_i^\star)\}_{i=1}^{5{,}000}\);
teacher \(\pi_{\mathrm T}\) (Qwen3.5-397B-A17B);
unmodified student initialization \(\pi_{\theta_0}\);
variant \(v\);
candidate-blind prompt \(p_1\);
trajectory-matching prompt \(p_2\).}\\
\multicolumn{2}{@{}p{\linewidth}@{}}{\textbf{Variants:}
\textcolor{jeftpink}{\textbf{JEFT Distillation}};
\textcolor{deftmcqblue}{\textbf{DEFT Distillation (Plan Only)}};
\textcolor{deftgenorange}{\textbf{DEFT Distillation (Full Interaction)}};
\textcolor{deftrlvrred}{\textbf{DEFT Distillation (Mixed Targets)}}.}\\
\multicolumn{2}{@{}p{\linewidth}@{}}{\textbf{Output:}
fine-tuned student \(\pi_\theta\).}\\[2pt]
1 & Initialize the supervised target set
    \(\mathcal T_v\leftarrow\varnothing\).\\
2 & \textcolor{deftrlvrred}{\textbf{If}
    \(v=\textsc{Mixed}\), fix an equal partition
    \(\mathcal D=\mathcal D_{\mathrm{plan}}\mathbin{\dot\cup}
    \mathcal D_{\mathrm{full}}\), with
    \(\lvert\mathcal D_{\mathrm{plan}}\rvert=
    \lvert\mathcal D_{\mathrm{full}}\rvert=2{,}500\).}\\
3 & \textbf{for each} \((X_i,\mathcal A_i,a_i^\star)\in\mathcal D\)
    \textbf{do}\\
4 & \deftvariantbox{jeftpink}{%
    \textcolor{jeftpink}{\textbf{JEFT Distillation:}}
    reveal \(\mathcal A_i\) together with \(X_i\) at the outset, query
    \(\pi_{\mathrm T}\) for one joint reasoning-and-selection response
    \(y_i^{\mathrm{joint}}\), and add the resulting single-turn example to
    \(\mathcal T_v\).}\\[2pt]
5 & \deftvariantbox{deftsharedgray}{%
    \textbf{All DEFT distillation variants:}
    query \(\pi_{\mathrm T}\) with \(p_1(X_i)\), without
    \(\mathcal A_i\), to obtain the candidate-blind plan \(y_{1,i}\).}\\[2pt]
6 & \deftvariantbox{deftmcqblue}{%
    \textcolor{deftmcqblue}{\textbf{DEFT Distillation (Plan Only):}}
    retain only the Turn-1 pair \((p_1(X_i),y_{1,i})\) in
    \(\mathcal T_v\); do not include candidate matching in the target.}\\[2pt]
7 & \deftvariantbox{deftgenorange}{%
    \textcolor{deftgenorange}{\textbf{DEFT Distillation (Full Interaction):}}
    after \(y_{1,i}\) is complete, reveal \(\mathcal A_i\), query
    \(\pi_{\mathrm T}\) with \(p_2\) for the matching response \(y_{2,i}\),
    and add the complete plan-then-match interaction to \(\mathcal T_v\).}\\[2pt]
8 & \deftvariantbox{deftrlvrred}{%
    \textcolor{deftrlvrred}{\textbf{DEFT Distillation (Mixed Targets):}}
    if \(i\in\mathcal D_{\mathrm{plan}}\), retain only \(y_{1,i}\) as in
    Plan Only; otherwise reveal \(\mathcal A_i\), generate \(y_{2,i}\), and
    retain the complete interaction as in Full Interaction.}\\[2pt]
9 & \textbf{end for}\\
10 & Initialize \(\pi_\theta\leftarrow\pi_{\theta_0}\) and perform supervised
     fine-tuning on \(\mathcal T_v\), freezing the visual encoder and updating
     the language model under the shared optimization configuration.\\
11 & \textbf{return} \(\pi_\theta\).\\
\end{tabular}
\end{tcolorbox}
}


\section{Causal Faithfulness under Future-Trajectory Exposure}
\label{app:motivation-study}

We provide the study design and detailed analysis for the comparison summarized in Section~\ref{sec:motivation}.
We examine whether revealing the logged GT future trajectory helps a teacher infer a faithful driving rationale or merely makes an already known outcome easier to justify.

\subsection{Study Design}

We evaluate the effect of GT-conditioned annotation on 100 strong-causal driving scenes,
including 70 Waymo scenes and 30 internal scenes.
We select scenes in which the logged future trajectory substantially deviates from constant-velocity extrapolation,
covering hard braking,
stopping from motion,
and sharp turns.
We detail the construction of the hard causal evaluation set in Appendix~\ref{app:hard-causal-mcq}.
For each scene,
we construct paired annotations using the same teacher.
Across the two experimental settings,
we fix Qwen3.5-397B-A17B as the teacher,
the same 12 visual frames,
ego-state and navigation text,
the complete system prompt and all user instructions outside the intervention
block,
the four-item causal-reasoning body,
the \texttt{HIGH\_LEVEL\_DECISION} output contract,
temperature \(0\),
and disabled thinking.
The two arms use a byte-identical system prompt and an otherwise identical
user template.
The sole intervention is one contiguous block containing the logged GT future
as 10 raw ego-frame waypoints:
the \emph{causal-planning} arm omits this block,
whereas the \emph{GT-conditioned} arm inserts it before the shared reasoning
instructions.
Both arms still ask the teacher to infer and commit to a high-level decision
from the scene; the GT-conditioned prompt does not ask the teacher to justify
a known action, repeat an action label, or select among candidates.
Consequently,
the paired comparison changes only the availability of future-trajectory
information while holding the task wording,
inputs,
model,
decoding,
and response format fixed.
The complete role-separated templates below expose the single insertion point
directly.

\definecolor{motpromptblue}{HTML}{0070C0}
\definecolor{motpromptpink}{HTML}{9A3F68}
\definecolor{motpromptborder}{HTML}{E8E8E8}

\newcommand{\motprompttitle}[1]{%
  {\color{motpromptblue}\bfseries #1}\par\vspace{2pt}}
\newcommand{\motpromptlabel}[1]{%
  {\color{motpromptblue}\bfseries #1}}
\newcommand{\motpromptvar}[1]{\texttt{\char`\{#1\char`\}}}
\newcommand{\motpromptdivider}{%
  \par\medskip{\color{motpromptborder}\hrule height 0.5pt}\medskip}
\newtcolorbox{motivationpromptbox}[2][]{%
  breakable,
  colback=motpromptblue!3!white,
  colframe=motpromptblue!75!black,
  title={#2},
  fonttitle=\bfseries\small,
  fontupper=\footnotesize,
  boxrule=0.8pt,
  arc=1.5mm,
  left=1.5mm,
  right=1.5mm,
  top=1.2mm,
  bottom=1.2mm,
  before skip=5pt,
  after skip=4pt,
  #1
}
\newcommand{\motivationfigurecaption}[2]{%
  \par\vspace{3pt}
  \refstepcounter{figure}\label{#1}%
  \noindent\hfill\begin{minipage}{0.96\linewidth}
  \small\textbf{Figure~\thefigure:} #2
  \end{minipage}\hfill\mbox{}\par\vspace{6pt}
}

\newcommand{\motivationpairedsystem}{%
You are an expert autonomous-driving planner. You are shown the recent
multi-view camera history and the ego vehicle's state. Reason causally about
the scene: connect what you observe to the maneuver through explicit
cause-and-effect, not a flat list of observations. Reason only from what is
provided to you, and commit to a single high-level driving decision for the
ego vehicle's 5-second future.\par
}

\newcommand{\motivationpaireduser}[1]{%
\motpromptlabel{Visual input:}
\motpromptvar{MULTI\_VIEW\_FRAMES, OLDEST TO NEWEST}\par
\motpromptlabel{Ego-state text:} \motpromptvar{EGO\_STATE}\par
\vspace{2pt}
Trajectory coordinates are in meters in the ego frame: +x is forward, +y is
to the left. Plan the ego's 5-second future motion
(t = 0.5s .. 5.0s).\par
#1
Reason about the scene and commit to a single high-level driving decision for
the ego's 5-second future. In 1--2 concise sentences each:\par
\vspace{2pt}
\textbf{1. Evidence:} from the history-visible scene, identify only the few
facts that causally constrain the future motion --- the road/route structure
and any element whose state changes what the ego can safely, legally, or
feasibly do next.\par
\textbf{2. Causal chain:} for each constraint reason scene
\(\rightarrow\) consequence (what becomes unsafe, illegal, infeasible, or
off-route) \(\rightarrow\) the maneuver it forces. Treat the ego's CURRENT
motion (its speed and acceleration) as only the starting condition to be acted
upon, NEVER as evidence that the motion should continue unchanged: the plan
follows from the scene, not from the current velocity. Unless the scene
positively shows that continuing unchanged is safe and on-route, let the
constraints --- not the momentum --- decide the maneuver.\par
\textbf{3. SPEED first, and firmly:} decide whether to keep speed, slow, or
stop from the hazards alone (signals, a lead or stopped vehicle, a crossing
agent, a stop line, a tight curve or intersection). This decision holds
REGARDLESS of which way the road goes --- never default to holding speed just
because the path looks clear or because you are unsure where the road
leads.\par
\textbf{4. DIRECTION:} from the history-visible road geometry, route
instruction, and scene constraints, commit to a direction for the 5-second
plan (straight / left / right); if the evidence is genuinely ambiguous, name
the 1--2 plausible directions and pick the best-supported one, briefly noting
the uncertainty. Do not invent or compute hypothetical trajectories here.\par
\vspace{2pt}
Then commit to a high-level decision on its own line in EXACTLY this
format:\par
\texttt{HIGH\_LEVEL\_DECISION: <speed profile (e.g.\ decelerate to a stop /
hold $\sim$Xm/s / slow then proceed) + direction (straight / left / right);
one sentence, with the main cause>}
}

\begin{motivationpromptbox}{Causal-Planning Prompt (GT Future Hidden)}
\motprompttitle{System Prompt}
\motivationpairedsystem
\motpromptdivider
\motprompttitle{User Message}
\motivationpaireduser{}
\end{motivationpromptbox}
\motivationfigurecaption{fig:motivation-causal-planning-template}{Complete
causal-planning chat template. The teacher receives the scene history and ego
state but no logged future. Per-scene inputs are shown as variables.}

\begin{motivationpromptbox}[colback=motpromptpink!3!white,colframe=motpromptpink!75!black]{GT-Conditioned Prompt (GT Future Visible)}
\motprompttitle{System Prompt}
\motivationpairedsystem
\motpromptdivider
\motprompttitle{User Message}
\motivationpaireduser{%
\vspace{3pt}
\noindent\colorbox{motpromptpink!10!white}{%
\parbox{\dimexpr\linewidth-2\fboxsep\relax}{%
GROUND-TRUTH future trajectory of the ego vehicle, logged from the recorded
future (ego frame, +x forward, +y left, 10 waypoints at
t = 0.5..5.0s):
\motpromptvar{(x1,y1), ..., (x10,y10)}.}}\par
\vspace{3pt}
}
\end{motivationpromptbox}
\motivationfigurecaption{fig:motivation-gtconditioned-template}{Complete
GT-conditioned chat template. It is identical to
Figure~\ref{fig:motivation-causal-planning-template} except for the highlighted
raw-waypoint GT block; no derived action label or rationalization-specific
instruction is introduced.}

\subsection{Human Evaluation}

We conduct a human evaluation of the two experimental settings.
Across the 100 scenes and two settings, we obtain 200 CoTs in total.
We ask two annotators to independently score every CoT along four dimensions:
grounding (GND),
absence of hallucination (NO-HALL),
specificity (SPEC),
and causal coherence (COH).
Each dimension is scored on a three-point ordinal scale,
where 0 denotes a clear failure with a consequential error,
1 denotes partial satisfaction with an omission or minor error,
and 2 denotes full satisfaction without a substantive error.
We thus collect \(200 \times 4 \times 2 = 1{,}600\) dimension-level ratings.
For each CoT,
we average the two annotations on each dimension and sum the four averaged
scores to obtain a causal-faithfulness score (CFS) from 0 to 8.
The annotators work independently,
and disagreements are retained rather than resolved through discussion or
adjudication.

We instantiate the three-point scale with dimension-specific observable criteria.
For GND,
0 indicates that the stated rationale conflicts with or is unsupported by the visible scene,
1 indicates that it uses some relevant scene evidence but omits or misinterprets a non-critical cue,
and 2 indicates that its decision-relevant claims are supported by the observed scene.
For NO-HALL,
0 indicates a consequential fabricated object, event, traffic control, or interaction,
1 indicates an unsupported but non-critical detail,
and 2 indicates no unsupported factual claim.
We count a CoT as a severe hallucination for an annotator when its
NO-HALL score is 0;
we compute the reported severe-hallucination rate by averaging this binary indicator over
annotators and CoTs within each experimental setting.
For SPEC,
0 indicates a generic rationale that could apply to unrelated scenes,
1 indicates limited use of scene-specific actors or geometry,
and 2 indicates sufficient reference to the particular actors, spatial relations, and traffic context that determine the maneuver.
For COH,
0 indicates that the conclusion does not follow from the stated evidence or contains a major contradiction,
1 indicates a broadly plausible causal chain with a missing link or minor inconsistency,
and 2 indicates a complete and internally consistent connection from scene evidence to the proposed maneuver.

For dimension-level scoring,
we anonymize the experimental settings by replacing their names with random
identifiers and randomizing presentation order.
We present the scene and a single CoT in each item and never show its paired counterpart alongside it.
The annotators are not informed of our anchoring hypothesis or which setting produced an item,
and we withhold the GT future trajectory during scoring.
We use the same teacher model,
scene inputs,
reasoning template,
decoding settings,
and output format for the two settings, as specified in the study design above.

After completing the independent dimension-level scoring,
each annotator separately evaluates all 100 scene-matched CoT pairs.
For each scene,
we present the two CoTs together in randomized order without experimental-setting
labels,
and ask the annotator to select the better rationale or record a tie using the same
grounding,
absence-of-hallucination,
specificity,
and causal-coherence criteria defined above.
The two annotators therefore provide 200 independent pairwise judgments:
causal planning is preferred in 121 judgments (60.5\%),
GT-conditioned reasoning is preferred in 48 (24.0\%),
and 31 judgments (15.5\%) are ties.

\section{Trajectory Codebook Construction and Representation Analysis}
\label{app:traj-token-repr}

In this section, we present the trajectory-codebook construction and representation analysis underlying \textbf{AD-MCQ}.
Notably, the codebook serves a specific role in our benchmark:
\textbf{(1) the codebook is a purely kinematic aggregation of motion trajectories and is aggregated independently of scene observations;
(2) we use it only to retrieve waypoint candidates when constructing \textbf{AD-MCQ}.
The codebook itself and its indices are never exposed to the policy,
which receives only the corresponding decoded waypoint candidates.}
We first formalize this representation,
then describe the reconstruction evaluation setup and analyze how codebook size and data scale determine reconstruction fidelity and prototype utilization.

\subsection{Trajectory Representation and Quantization}

A logged future is represented as a $5\,\mathrm{s}$ ego-frame trajectory sampled at $2\,\mathrm{Hz}$,
$\mathbf P=(\mathbf p_1,\ldots,\mathbf p_T)\in\mathbb R^{T\times2}$ with $T=10$.
Flattening $\mathbf P$ yields a $20$-dimensional vector.
Given $N$ such trajectories,
we apply $K$-means to obtain
$\mathcal C=\{\mathbf C_1,\ldots,\mathbf C_K\}$,
where each centroid $\mathbf C_k\in\mathbb R^{T\times2}$ represents a complete speed and lateral-motion profile.
As defined in Eq.~\ref{eq:traj-token-quantization},
each trajectory is assigned to its nearest centroid under cumulative squared waypoint distance.

For implementation,
we associate each codebook index $k$ with a symbolic identifier
$\tau_k\equiv\texttt{<traj\_k>}$.
Encoding returns the identifier of the nearest prototype,
whereas decoding retrieves the complete waypoint sequence,
$\operatorname{Dec}(\tau_k)=\mathbf C_k$.
These identifiers provide a discrete indexing space for candidate construction;
\textbf{AD-MCQ} serializes the decoded coordinates rather than the identifiers themselves.
Consequently,
quantization remains external to the VLM and does not require modifying its vocabulary or generating dense coordinates autoregressively.

\subsection{Reconstruction Evaluation Setup}

We pool approximately $5.1{\times}10^{5}$ logged trajectories from the Waymo Open E2E corpus and an internal driving corpus.
Waymo ego \texttt{future\_states} are subsampled from $4\,\mathrm{Hz}$ to $2\,\mathrm{Hz}$;
both sources use the same $5\,\mathrm{s}$,
$10$-waypoint,
ego-frame,
meter convention and are therefore clustered jointly.
We reserve $5\%$ ($25{,}739$ trajectories) from codebook fitting for same-distribution out-of-sample evaluation and use an independent Waymo validation split ($106{,}360$ trajectories) for cross-source evaluation.

Clustering uses MiniBatchKMeans with batch size $10{,}000$,
$300$ iterations,
and three initializations.
For each $(N,K)$ configuration,
we repeat K-means with three initialization seeds and report the mean over seeds separately for the fitting corpus,
the held-out split,
and the independent Waymo split.
We measure average displacement error (ADE),
final displacement error (FDE),
their p50/p95/p99 tail statistics,
and codebook utilization,
defined as the fraction of prototypes assigned at least one evaluation trajectory.

\subsection{Resolution as the Codebook Scales}

Table~\ref{tab:traj-k-res} reports the resolution sweep at the full clustering-set size.
Increasing $K$ reduces ADE and FDE smoothly rather than producing a sharp saturation point;
each doubling lowers in-sample ADE by approximately $15\%$.
The improvement follows the approximate trend $\mathrm{ADE}\approx C K^{-0.25}$,
but increasingly fine codebooks allocate prototypes to sparse motions that are not recovered on held-out trajectories.

\begin{table}[h]
\centering
\begin{tabular}{llllllll}
\hline
$K$ & In ADE & Out ADE & Cross ADE & In FDE & Out FDE & In Util. & Out Util. \\
\hline
256   & 0.680 & 0.675 & 0.704 & 1.281 & 1.273 & 100.0\% & 100.0\% \\
512   & 0.567 & 0.563 & 0.592 & 1.063 & 1.057 & 100.0\% & 100.0\% \\
1024  & 0.480 & 0.478 & 0.505 & 0.890 & 0.887 & 100.0\% & 100.0\% \\
2048  & 0.399 & 0.399 & 0.432 & 0.729 & 0.729 & 100.0\% & 100.0\% \\
4096  & 0.333 & 0.336 & 0.374 & 0.601 & 0.605 & 100.0\% & 98.7\% \\
8192  & \textbf{0.283} & 0.290 & 0.331 & 0.504 & 0.516 & 99.7\% & 92.0\% \\
16384 & 0.242 & \textbf{0.252} & \textbf{0.297} & 0.426 & 0.446 & 99.4\% & 73.7\% \\
\hline
\end{tabular}
\caption{Trajectory-codebook resolution at $N{=}489{,}042$ clustering trajectories.
Errors are measured in meters and per-cell ADE standard deviation is at most $0.005$.
Larger codebooks improve reconstruction fidelity but reduce out-of-sample utilization beyond $K{=}8192$.}
\label{tab:traj-k-res}
\end{table}

\begin{table}[h]
\centering
\begin{tabular}{@{}llllll@{}}
\toprule
\textbf{$K$} & \textbf{$N$} & \textbf{In ADE} & \textbf{Out ADE}
& \textbf{Gap} & \textbf{Out Util.} \\
\midrule
\multirow{7}{*}{$1024$}
& 5{,}000     & 0.407 & 0.528 & $+0.121$ & 99.8\% \\
& 10{,}000    & 0.442 & 0.509 & $+0.067$ & 99.8\% \\
& 20{,}000    & 0.464 & 0.496 & $+0.032$ & 100.0\% \\
& 50{,}000    & 0.476 & 0.489 & $+0.013$ & 100.0\% \\
& 100{,}000   & 0.476 & 0.479 & $+0.003$ & 100.0\% \\
& 200{,}000   & 0.477 & 0.477 & $\sim0$   & 100.0\% \\
& 489{,}042   & 0.480 & 0.478 & $\sim0$   & 100.0\% \\
\midrule
\multirow{4}{*}{$8192$}
& 50{,}000    & 0.243 & 0.303 & $+0.060$ & 89.6\% \\
& 100{,}000   & 0.266 & 0.297 & $+0.031$ & 90.4\% \\
& 200{,}000   & 0.274 & 0.290 & $+0.016$ & 91.6\% \\
& 489{,}042   & 0.283 & 0.290 & $+0.007$ & 92.0\% \\
\bottomrule
\end{tabular}
\caption{Reconstruction generalization as the clustering corpus grows.
At fixed $K$,
additional trajectories reduce the out-of-sample gap and stabilize prototype utilization.}
\label{tab:traj-n-sat}
\end{table}

The resolution--coverage trade-off is visible in both Table~\ref{tab:traj-k-res} and Figure~\ref{fig:traj-token-representation}a.
Out-of-sample utilization remains complete through $K{=}2048$,
is $98.7\%$ at $K{=}4096$ and $92.0\%$ at $K{=}8192$,
but falls to $73.7\%$ at $K{=}16384$.
The independent Waymo split exhibits the same monotonic resolution trend with a consistent $0.02$--$0.05\,\mathrm{m}$ ADE offset,
showing that the comparison across $K$ is not specific to a single held-out split.

\subsection{Data Scale and Generalization}

We next vary the number of clustering trajectories while fixing $K$.
As summarized in Table~\ref{tab:traj-n-sat} and Figure~\ref{fig:traj-token-representation}b,
small clustering sets yield an artificially low in-sample error but a larger held-out error because their centroids specialize to incidental sample positions.
Increasing $N$ closes this gap,
with in-sample and out-of-sample reconstruction approaching convergence once the number of trajectories per prototype becomes sufficiently large.

For $K{=}1024$,
the gap is effectively closed once $N$ reaches approximately $10^{5}$;
for $K{=}8192$,
it decreases to $0.007\,\mathrm{m}$ at the full data scale.
The long tail remains the main source of quantization error:
at full data,
ADE p99 decreases from $2.74\,\mathrm{m}$ at $K{=}256$ to $1.14\,\mathrm{m}$ at $K{=}16384$,
and FDE p99 decreases from $5.82\,\mathrm{m}$ to $2.33\,\mathrm{m}$.
Thus,
larger codebooks improve rare-motion reconstruction but do not eliminate long-tail error by themselves.

\subsection{Codebook Selection}

Our choice of $K{=}8192$ follows from the joint behavior of reconstruction fidelity,
held-out utilization,
and data support.
At the full clustering scale,
it achieves $0.290\,\mathrm{m}$ out-of-sample ADE and $0.516\,\mathrm{m}$ out-of-sample FDE while retaining $92.0\%$ utilization.
Doubling the codebook further improves displacement error,
but the fraction of prototypes exercised out of sample drops by more than $18\%$.
Conversely,
smaller codebooks retain nearly complete utilization but provide coarser trajectory distinctions.
We therefore use $K{=}8192$ throughout \textbf{AD-MCQ} as the operating point that preserves fine-grained speed and lateral-motion patterns without allocating a large fraction of the codebook to unsupported prototypes.

\section{Detailed Experimental Settings}
\label{app:experimental-details}

\subsection{Data Sources and AD-MCQ Construction}
\label{app:ad-mcq-construction}

We construct AD-MCQ from \(514{,}781\) scene--trajectory pairs drawn from Waymo Open E2E and an internal driving corpus.
For each scene, every front-left, front, and front-right camera stream is decoded into four historical frames sampled at \(2\,\mathrm{Hz}\), together with the ego state, navigation context, and logged future trajectory.
Qwen3-VL groups each adjacent pair of frames into one temporal patch, so the rendered prompt displays two temporal-patch timestamps (\texttt{<0.2 seconds><1.2 seconds>}) per camera stream even though the model input contains four decoded frames; Appendix~\ref{app:two-turn-policy-prompt} shows the resulting prompt representation.

We then construct the three scene-disjoint downstream splits in
Table~\ref{tab:ad-mcq-splits}.
Training follows the natural scene distribution,
whereas Dev and Test emphasize causally difficult long-tail maneuvers.
Following Section~\ref{sec:mcq},
each question contains $M{=}6$ decoded options retrieved from a $K{=}8192$-prototype codebook.
Every option is a $5\,$s ego-frame trajectory with $10$ waypoints sampled at $2\,$Hz.
The decoded candidates are logged-data-derived waypoint prototypes.

\paragraph{Notation and Precomputation.}
We write \(X_i=(V_i,H_i,I_i)\) for the visual history, ego history and
current motion state, and navigation instruction of scene \(i\).
This factorization preserves three complementary signals required for driving:
\(V_i\) provides spatial coverage and short-term temporal evidence about scene
dynamics, \(H_i\) supplies the ego-motion context needed to interpret those
observations, and \(I_i\) specifies route-level intent when multiple futures
are geometrically feasible. It also follows the established VLA input
interface of multi-view, multi-frame images, ego-vehicle states, and
high-level navigation instructions~\cite{zhou2026autovla}.
For split \(s\), Algorithm~\ref{alg:ad-mcq-construction} maps the input
scene--trajectory pairs to
\(\mathcal D_s=\{(X_i,\mathcal A_i,a_i^\star)\}\), where
\(\mathcal A_i\) is the shuffled six-option set and \(a_i^\star\) is the
position of the quantized logged future after shuffling.

We precompute one \(K\times K\) similarity matrix for the entire codebook and
reuse it for every scene and split. Specifically, \(d_{\min}\) and
\(d_{\max}\) in Eq.~\ref{eq:traj-token-similarity} are the minimum and maximum
over all \(K^2\) pairwise trajectory ADE values, rather than statistics of an
individual split or candidate pool. We clip the resulting similarities to
\([0,1]\) and set the diagonal to one. Thus, candidate sampling only indexes
the fixed row associated with the oracle token; it does not renormalize
similarities per instance.

\begin{table}[h]
\centering
\small
\setlength{\tabcolsep}{3pt}
\begin{tabular}{llll}
\toprule
\textbf{Split} & \textbf{Num} & \textbf{Scene Preference} & \textbf{Distractors} \\
\midrule
Train & $5{,}000$ & Natural distribution. & Random \\
\midrule
Dev & $100$ & $55$ stop-from-motion; $30$ hard brakes; $15$ sharp turns. & Structured \\
\midrule
Test & $500$ & $250$ straight/brake/stop; $125$ left; $125$ right. & Structured \\
\bottomrule
\end{tabular}
\caption{Split-specific \textbf{AD-MCQ} distractors:
$5$ random hard negatives for Train;
$2$ scale-matched $+$ $1$ constant-velocity $+$ $2$ hard negatives for
Dev/Test.}
\label{tab:ad-mcq-splits}
\end{table}

\paragraph{Training Split: Scene and Candidate Construction.}
We randomly sample $5{,}000$ training scenes from the natural scene distribution.
For each scene,
we instantiate the oracle with the nearest codebook prototype to the logged future
and sample five distinct distractors at random from the hard-negative pool in
Eq.~\ref{eq:mcq-hard-pool}.
For the main experiments,
this pool uses $\rho_{\min}{=}0.30$ and $\rho_{\max}{=}0.85$.
We sample uniformly without replacement within this band using a deterministic
per-record random-number generator whose seed is derived from the base seed and
record key as \(\operatorname{SHA1}(\texttt{base\_seed:record\_key})\).
Consequently, repeated construction with the same base seed produces the same
candidates.
Sampling candidates broadly within this difficulty range avoids teaching the
policy a fixed distractor template.

\paragraph{Dev and Test: Hard-Causal Scene Construction.}
\label{app:hard-causal-mcq}

Dev and Test evaluate whether a model can identify the scene evidence that causally supports a driving decision.
We apply the ordered classifier in Table~\ref{tab:hard-causal-rules}.
Let $v_0$ and $v_f$ denote the current and final logged speeds,
$L_{\mathrm{gt}}$ the logged-future path length,
$L_{\mathrm{cv}}$ the path length under constant-velocity extrapolation,
and $\Delta\psi$ the net heading change of the logged future.
We define the braking ratio as
$b=1-L_{\mathrm{gt}}/L_{\mathrm{cv}}$.
The minimum current-speed threshold is $v_{\min}{=}3.0\,\mathrm{m/s}$ for
Waymo and $1.5\,\mathrm{m/s}$ for the internal corpus.
Scenes classified as low-speed or routine straight driving are discarded;
the remaining stop-from-motion, hard-braking, and sharp-turn scenes form the
hard-causal pool.
For ranking within each retained class,
we use
$g_{\mathrm{cv}}=\lVert p_T^{\mathrm{cv}}-p_T^{\mathrm{gt}}\rVert_2$,
the endpoint gap between constant-velocity extrapolation and the logged future,
and retain up to the top $800$ scenes per upstream pool and class.
Each resulting question includes the constant-velocity trajectory as an
explicit distractor. We retain only instances for which its quantized token is
distinct from the oracle and belongs to the evaluation hard-negative pool
defined below.

\begin{table}[h]
\centering
\small
\begin{tabular}{lll}
\toprule
\textbf{Ordered Class} & \textbf{Condition} & \textbf{Disposition} \\
\midrule
Low speed & $v_0<v_{\min}$ & Discard \\
Stop from motion & $v_f<0.5\,\mathrm{m/s}$ and $b>0.55$ & Retain \\
Hard brake & $b>0.45$ and $v_f<0.6v_0$ & Retain \\
Sharp turn & $\lvert\Delta\psi\rvert>25^\circ$ & Retain \\
Routine straight & Otherwise & Discard \\
\bottomrule
\end{tabular}
\caption{Ordered hard-causal scene classifier. The first satisfied condition
determines the class.}
\label{tab:hard-causal-rules}
\end{table}

For Dev,
we manually review the candidate pool and retain the most extreme $55$ stop-from-motion,
$30$ hard-braking,
and $15$ sharp-turn cases.
Test broadens directional coverage with $250$ straight braking or stopping scenes,
$125$ left-turn scenes,
and $125$ right-turn scenes;
the turns are primarily sharp.
The median gap between the logged future and constant-velocity extrapolation is $28.5\,$m on Dev and $27.0\,$m on Test,
confirming that the larger test set preserves the long-tail focus.
Dev is used for the motivation study and reward development;
the scene-disjoint \textbf{AD-MCQ-500} Test split is used only for final evaluation.

\paragraph{Dev and Test: Structured Candidate Construction.}
For each Dev or Test scene,
we instantiate the oracle with the nearest codebook prototype to the logged future
and construct five distractors as a structured mixture.
All five distractors come from the split-specific evaluation hard-negative
pool
\begin{equation}
    \mathcal H_i^{\mathrm{eval}}
    =
    \left\{z\in[K]\setminus\{z_i^\star\}:
    0.30\leq\rho_{z,z_i^\star}\leq0.92\right\}.
    \label{eq:eval-hard-negative-pool}
\end{equation}
We select two scale-matched hard negatives;
they match the oracle's endpoint-displacement scale while differing in
trajectory shape, countering shortcuts based only on displacement magnitude.
For prototype \(k\), define its scale as
\begin{equation}
    q_k=\lVert\mathbf c_{k,T}\rVert_2.
    \label{eq:trajectory-endpoint-scale}
\end{equation}
We define
\(\operatorname{ScaleMatch}_{0.15}(z,z_i^\star)\) by
\begin{equation}
    \lvert q_z-q_{z_i^\star}\rvert
    \leq 0.15\,q_{z_i^\star}.
    \label{eq:scale-match}
\end{equation}
Among unused prototypes in \(\mathcal H_i^{\mathrm{eval}}\) satisfying this
constraint, we take the two with the smallest
\(\lvert q_z-q_{z_i^\star}\rvert\). Thus, ``scale'' denotes the Euclidean
displacement of the final \(5\)-s waypoint from the current ego origin, not
path length or speed.

For the constant-velocity candidate, we use the recorded planar ego velocity
\(\mathbf v_i=\texttt{record["velocity"]}\allowbreak\texttt{[:2]}\), rather than estimating
velocity from the sampled ego-history positions. With
\(\Delta t=0.5\,\mathrm{s}\), we construct
\begin{equation}
    \mathbf p_{i,t}^{\mathrm{cv}}=t\Delta t\,\mathbf v_i,
    \qquad t=1,\ldots,10,
    \label{eq:constant-velocity-candidate}
\end{equation}
and retrieve its nearest codebook prototype using
Eq.~\ref{eq:traj-token-quantization},
countering momentum-based extrapolation.
For every retained instance, this prototype is distinct from the oracle and
the scale-matched candidates and lies in
\(\mathcal H_i^{\mathrm{eval}}\).
Two additional distinct samples satisfying
$0.30\leq\rho\leq0.85$ complete the five distractors.
All random draws from this band are uniform without replacement.

\begin{table}[t]
\centering
\small
\setlength{\tabcolsep}{2pt}
\begin{tabular}{@{}m{0.22\columnwidth}m{0.20\columnwidth}m{0.54\columnwidth}@{}}
\toprule
\textbf{Candidate Component} & \textbf{Count} & \textbf{Eligibility Rule} \\
\midrule
Oracle & $1$ & Nearest codebook prototype to the logged future. \\
\midrule
Scale-matched & $2$ & In \(\mathcal H_i^{\mathrm{eval}}\), with
endpoint-displacement mismatch at most $0.15$ relative to the oracle;
smallest mismatch first. \\
\midrule
Constant velocity & $1$ & In \(\mathcal H_i^{\mathrm{eval}}\), nearest to
\(\mathbf p_t^{\mathrm{cv}}=t(0.5\,\mathrm{s})\mathbf v_i\). \\
\midrule
General hard negative & $2$ & Distinct prototypes satisfying
$0.30\leq\rho\leq0.85$. \\
\bottomrule
\end{tabular}
\caption{Structured six-candidate construction used for Dev and Test.
The oracle and previously selected prototypes are excluded during each
distractor-selection step.}
\label{tab:structured-candidate-rules}
\end{table}

\paragraph{Candidate Validity and Deduplication.}
We maintain an exclusion set containing the oracle and every selected
distractor, so scale-matched, constant-velocity, and hard-negative candidates
cannot duplicate one another. Every retained Train, Dev, and Test instance has
enough eligible prototypes to obtain the required five distinct distractors;
no sampling fallback outside its split-specific hard-negative pool is used.
We randomly shuffle the oracle and five distractors and record \(a_i^\star\)
as the oracle's shuffled position.

The positive option is consistent with the logged future,
route intent,
and map constraints,
whereas negative options remain plausible explicit futures.
When possible,
distractors match the positive option in displacement,
speed range,
endpoint distance,
or temporal horizon but differ in lane choice,
yielding behavior,
braking timing,
obstacle clearance,
or route compliance.
We randomize option order to reduce position bias.
\textbf{This split-specific construction deliberately tests whether behavior
learned from diverse random negatives transfers to targeted
endpoint-displacement- and momentum-based distractors.}

Algorithm~\ref{alg:ad-mcq-construction} summarizes how we convert each
scene--trajectory pair into an \textbf{AD-MCQ} instance. The split-specific
sampling operator follows the settings in
Appendix~\ref{app:ad-mcq-construction}: Train uses five samples from the broad
hard-negative pool, whereas Dev and Test use two scale-matched
negatives, one constant-velocity negative, and two additional hard negatives.
In the
algorithm, \(\operatorname{ScaleMatch}_{0.15}\) compares the endpoint
displacement magnitudes of two trajectory prototypes, as defined in
Appendix~\ref{app:ad-mcq-construction}.

\refstepcounter{deftalgorithm}
\label{alg:ad-mcq-construction}
\begin{tcolorbox}[
  colback=white,
  colframe=black!55,
  boxrule=0.6pt,
  arc=0pt,
  left=6pt,
  right=6pt,
  top=6pt,
  bottom=6pt
]
\noindent\textbf{Algorithm \thedeftalgorithm: Split-Specific AD-MCQ Construction}
\par\smallskip\hrule\smallskip
\small
\begin{tabular}{@{}r@{\hspace{0.7em}}p{\dimexpr\linewidth-2.5em\relax}@{}}
\multicolumn{2}{@{}p{\linewidth}@{}}{\textbf{Input:}
scene--trajectory pairs
\(\mathcal S=\{(X_i,\mathbf P_i^{\mathrm{gt}})\}\) assigned to split
\(s\in\{\mathrm{Train},\mathrm{Dev},\mathrm{Test}\}\);
codebook \(\mathcal C=\{\mathbf C_k\}_{k=1}^{K}\);
option count \(M=6\);
recorded planar ego velocity \(\mathbf v_i\);
similarities \(\rho\) from Eq.~\ref{eq:traj-token-similarity}.}\\
\multicolumn{2}{@{}p{\linewidth}@{}}{\textbf{Output:}
\(\mathcal D_s=\{(X_i,\mathcal A_i,a_i^\star)\}\).}\\[2pt]
1 & Initialize \(\mathcal D_s\leftarrow\varnothing\).\\
2 & \textbf{for each} \((X_i,\mathbf P_i^{\mathrm{gt}})\in\mathcal S\)
    \textbf{do}\\
3 & \hspace*{1em}Quantize the logged future:
    \(z_i^\star\leftarrow
    \arg\min_{k\in[K]}\sum_{t=1}^{T}
    \lVert\mathbf p_{i,t}^{\mathrm{gt}}-\mathbf c_{k,t}\rVert_2^2\).\\
4 & \hspace*{1em}Initialize the selected distractors
    \(\mathcal N_i\leftarrow\varnothing\).\\
5 & \hspace*{1em}\textbf{if} \(s=\mathrm{Train}\) \textbf{then}\\
6 & \hspace*{2em}Form
    \(\mathcal H_i^{\mathrm{tr}}=
    \{z\ne z_i^\star:0.30\leq\rho_{z,z_i^\star}\leq0.85\}\)
    and uniformly sample \(M-1\) distinct indices into
    \(\mathcal N_i\).\\
7 & \hspace*{1em}\textbf{else} \(\;(s\in\{\mathrm{Dev},\mathrm{Test}\})\)\\
8 & \hspace*{2em}Form
    \(\mathcal H_i^{\mathrm{eval}}=
    \{z\ne z_i^\star:0.30\leq\rho_{z,z_i^\star}\leq0.92\}\);
    add the two indices in this pool satisfying
    \(\operatorname{ScaleMatch}_{0.15}(z,z_i^\star)\) with the
    smallest endpoint-scale mismatches.\\
9 & \hspace*{2em}Form
    \(\mathbf p_{i,t}^{\mathrm{cv}}=t(0.5\,\mathrm{s})\mathbf v_i\)
    for \(t=1,\ldots,10\), quantize it as
    \(z_i^{\mathrm{cv}}\leftarrow z(\mathbf P_i^{\mathrm{cv}})\), and
    add \(z_i^{\mathrm{cv}}\in\mathcal H_i^{\mathrm{eval}}\), which is
    distinct from the two scale-matched indices.\\
10 & \hspace*{2em}Uniformly sample two additional unused indices in
    \(\mathcal H_i^{\mathrm{eval}}\) satisfying
    \(\rho_{z,z_i^\star}\leq0.85\).\\
11 & \hspace*{1em}\textbf{end if}\\
12 & \hspace*{1em}Randomly permute
    \(\{z_i^\star\}\cup\mathcal N_i\) to obtain
    \((z_{i,1},\ldots,z_{i,M})\), and record the oracle position
    \(a_i^\star\) such that \(z_{i,a_i^\star}=z_i^\star\).\\
13 & \hspace*{1em}Decode the indices into explicit waypoint options:
    \(\mathcal A_i\leftarrow
    (\mathbf C_{z_{i,1}},\ldots,\mathbf C_{z_{i,M}})\).\\
14 & \hspace*{1em}Add
    \((X_i,\mathcal A_i,a_i^\star)\) to \(\mathcal D_s\).\\
15 & \textbf{end for}\\
16 & \textbf{return} \(\mathcal D_s\).\\
\end{tabular}
\end{tcolorbox}

\begin{table}[t]
\centering
\begin{tabular}{llllll}
\toprule
\textbf{Quantity} & \textbf{Mean} & \textbf{P50} & \textbf{P75} &
\textbf{P90} & \textbf{P95} \\
\midrule
Oracle ADE (\(\mathrm{m}\))             & 0.45  & 0.227 & 0.483 & 0.933 & 1.630 \\
Oracle FDE (\(\mathrm{m}\))             & 0.79  & 0.319 & 0.684 & 1.576 & 2.671 \\
Oracle ADE / \(L_{\mathrm{gt}}\) (\%)    & 3.0   & 2.4   & 3.7   & 5.8   & 7.4   \\
All distractors ADE (\(\mathrm{m}\))    & 19.77 & 17.94 & 24.93 & 35.27 & 44.25 \\
Nearest distractor ADE (\(\mathrm{m}\)) & 7.16  & 7.05  & 8.73  & 10.27 & 11.86 \\
\bottomrule
\end{tabular}
\caption{Oracle reconstruction fidelity and candidate separation on
\textbf{AD-MCQ-500}. Distractor ADE is measured against the logged future;
the nearest distractor is selected independently for each instance.
\textbf{P50}, \textbf{P75}, \textbf{P90}, and \textbf{P95} denote the
50th, 75th, 90th, and 95th percentiles, respectively, of each quantity
across evaluation instances.}
\label{tab:ad-mcq-oracle-fidelity}
\end{table}

\paragraph{Oracle Fidelity and Candidate Separation.}
The oracle provides a high-fidelity representation of the logged motion,
while the alternative candidates constitute geometrically distinct plans
rather than duplicate quantizations or small coordinate perturbations.
Table~\ref{tab:ad-mcq-oracle-fidelity} supports this conclusion:
the oracle prototype accurately reconstructs the logged future for most
\textbf{AD-MCQ-500} instances, and its displacement error is typically small
relative to the scale of the trajectory space.
By contrast, even the nearest distractor in each instance remains
substantially farther from the logged future.
Its error exceeds the oracle error in all \(500\) instances, with a
median oracle--distractor gap of \(6.74\,\mathrm{m}\).
We do not assume that the logged future is the unique safe trajectory:
AD-MCQ operationalizes planning as discrimination among explicit plan
hypotheses relative to demonstrated behavior, not as an exhaustive
certification of every feasible future.
Nevertheless, the cross-domain gains in
Table~\ref{tab:nuscenes-generalization} and the consistent gains under
resampled candidate counts and similarity bands in
Figure~\ref{fig:candidate-set-robustness} and
Table~\ref{tab:candidate-set-ablation} show that \textbf{the learned capability
transfers across both driving domains and candidate constructions, rather
than merely recovering a fixed recorded action or exploiting one particular
distractor geometry.}

\subsection{Evaluation Settings and Benchmarks}
\label{app:evaluation-settings}

\paragraph{AD-Specific Evaluation.}
All AD-specific evaluations use a $65{,}536$-token context window and the following decoding configuration: $T{=}1.0$, top-$p{=}0.95$, thinking enabled, and a generation limit of $12{,}000$ tokens per turn for \textbf{DEFT} inference and $24{,}576$ tokens in total.
For \textbf{JEFT}, we directly set the token budget to 24,576.
We evaluate candidate selection on \textbf{AD-MCQ-500} using strict option accuracy.
Each reported model is evaluated eight times under the same configuration,
and we report the mean over these runs.
On the same $500$ scenes,
we assess candidate-blind Turn-1 outputs with two complementary metrics.
\textbf{Normed-CFS} (Normalized Causal-Faithfulness Score) is a GT-blind
automatic score over the same four dimensions used in the human study:
grounding, absence of hallucination, specificity, and causal coherence.
The judge assigns a binary value
$b_d\in\{0,1\}$ to each dimension
$d\in\{\mathrm{GND},\mathrm{NO\text{-}HALL},\mathrm{SPEC},\mathrm{COH}\}$,
and we compute
\(\mathrm{Normed\text{-}CFS}=\frac{1}{4}\sum_d b_d\in[0,1]\).
This differs deliberately from the $0/1/2$ ordinal scale used by human
annotators in Section~\ref{sec:motivation}.
Human annotators can reliably distinguish partial from full satisfaction,
whereas this intermediate category is less stable for a model judge;
we therefore ask the automatic evaluator only for mechanically defined binary
decisions~\cite{hashemi2024llm,rao2026autorubric}.
We report this normalized automatic score as \textbf{CFS} in the result tables.
\textbf{HLD} (High-Level-Decision Consistency) measures agreement between
the predicted high-level decision and the GT trajectory, requiring both direction
and speed to match the oracle action induced by the GT waypoints.
Qwen3.5-397B-A17B judges these two open-ended metrics;
strict candidate accuracy uses exact option matching and no LLM judge.
The complete CFS and HLD evaluation-judge prompts are provided in
Appendix~\ref{app:open-eval-judge-prompts}.

\paragraph{General-Capability Evaluation.}
To measure capability retention,
we evaluate $12$ benchmarks with VLMEvalKit and report four category averages:
basic visual perception,
embodied spatial reasoning,
3D/multi-view reasoning,
and RefSpatial grounding.
The category averages are computed over the following benchmark triplets:
\textbf{Basic Visual} comprises CV-Bench-2D, CV-Bench-3D, and DA-2K;
\textbf{Embodied Spatial} comprises EmbSpatialBench, RoboSpatialHome, and ERQA;
\textbf{3D/Multi-View} comprises 3DSRBench, MMSIBench, and ViewSpatialBench;
and \textbf{RefSpatial} comprises the Location, Placement, and Unseen splits of
RefSpatial-Bench.
Their macro-average is reported as \textbf{General AVG}.
All scores are percentages obtained with greedy decoding and without a
chain-of-thought prompt, using the same configuration across main-table
evaluations.

\paragraph{Cross-Domain AD Evaluation.}
We construct an external $500$-scene set from nuScenes val using the same trajectory codebook,
six-candidate construction,
ego-motion inputs,
the two-turn deferred-exposure interface,
and eight independently sampled evaluation runs, matching \textbf{AD-MCQ-500}.
The resulting evaluation preserves the task and output contract while changing the driving domain.
We use Qwen3.5-397B-A17B to judge the two external-set open metrics,
matching the evaluator used for the in-domain evaluation.

\subsection{Models, Distillation, and RLVR Optimization}
\label{app:distillation-rlvr-optimization}

\paragraph{Models.}
We use Qwen3-VL-8B-Instruct and Qwen3.5-4B as the base policies.
Qwen3.5-397B-A17B~\cite{qwen3.5} supplies CoT annotations,
and Qwen3.6-35B-A3B serves as the offline vision-language rubric generator and online text-only rubric grader.
The cold-start SFT configuration is detailed in
Appendix~\ref{app:cold-start-sft}.

\paragraph{Controlled Distillation Baselines.}
\textbf{DEFT Distillation (Plan Only)},
\textbf{DEFT Distillation (Full Interaction)},
and \textbf{DEFT Distillation (Mixed Targets)} each use $5{,}000$ teacher-labeled scenes and start from the unmodified Qwen3-VL-8B-Instruct policy.
Their targets are respectively all Turn-1-only,
all complete two-turn,
or a fixed $2{,}500/2{,}500$ split of the two formats.
All three runs use TP${=}2$,
global batch size $32$,
micro-batch size $1$,
maximum sequence length $20{,}480$,
and a constant learning rate of $1\!\times\!10^{-5}$.
The visual encoder is frozen and the language model is updated.

\DistillationTrainingAlgorithm

\paragraph{RLVR Optimization.}
We optimize the complete two-turn sequence with GRPO and one trajectory-level scalar advantage shared by the generated tokens of both turns.
For each optimizer step we sample $16$ rollouts for each of $64$ questions,
giving $1{,}024$ trajectories per step.
The actor uses a constant learning rate of $1\!\times\!10^{-6}$ without warmup.
The policy loss uses token-mean aggregation and gradient clipping at $1.0$,
with the normalized advantage clipped to $[-3.75,3.75]$.
A KL loss with coefficient $0.001$ regularizes the policy but is not added to the reward.
Training is on-policy (one update per sampled batch) and uses no critic or entropy bonus.
Rollouts are sampled with $T{=}1.0$, top-$p{=}0.95$, and top-$k{=}-1$.
We use prompt and response limits of $12{,}288$ and $24{,}576$ tokens,
respectively,
with a $12{,}000$-token generation limit per turn and a $65{,}536$-token
rollout context limit.
Rollout inference uses a device-memory utilization of $0.9$.

The Qwen3-VL-8B and Qwen3.5-4B experiments use the same optimization configuration.
All main RLVR runs start directly from their respective base models.

\paragraph{Reward and Grader Configuration.}
The exact verifier parses the final occurrence of \texttt{FINAL\_CHOICE: [A--F]} and assigns one only when it matches the shuffled oracle option; malformed and incorrect responses receive zero.
For \textbf{DEFT-RLVR},
each question has $6$--$10$ positive atomic rubric criteria with integer weights in $[1,10]$.
Qwen3.6-35B-A3B generates these criteria offline with temperature $0.3$,
top-$p=0.9$,
a $1{,}024$-token output limit,
and thinking disabled.
The same model grades each correct normalized Turn-1 trace online without images,
candidate trajectories,
the oracle,
or answer letters,
using temperature $1.0$,
top-$p=0.7$,
a $256$-token output limit,
and thinking disabled.
It is served in bf16 with tensor parallelism $16$ and a $65{,}536$-token context window.
If grading fails after retry handling,
we fall back to the exact-correctness reward;
an incorrect final choice always receives zero.

Before either rubric-based grader is called,
we isolate the candidate-blind Turn-1 trace by truncating the serialized
interaction at the injected Part-2 option block and removing any residual
injected option text or Turn-2 response.
We then remove reasoning-wrapper tags,
the \texttt{HIGH\_LEVEL\_DECISION} line,
and structural headers such as \texttt{PART}, \texttt{REASONING}, and
\texttt{===}, while retaining the substantive evidence and causal reasoning.
We denote the resulting trace by
\(\widetilde y_{1,i,j}:=\operatorname{Normalize}(y_{1,i,j})\).
This normalization prevents the grader from using the committed maneuver,
candidate options, or final answer as a proxy for reasoning quality.

Controlled SFT comparisons use equal numbers of teacher-annotated examples.
Controlled RLVR comparisons share the base VLM,
training scenes,
prompts,
rollout budget,
and core optimization settings unless explicitly stated otherwise.

\subsection{Baselines and Controlled Variants}
\label{app:baselines-controlled-variants}

We compare \textbf{DEFT-RLVR} against the following task-matched baselines and
controlled DEFT variants,
all using the same AD-MCQ task and candidate representation.
\begin{itemize}
    \item \textbf{Base VLM (Direct MCQ).}
    We run the base model in a single turn with candidates visible
    from the outset and evaluate only its exact-choice accuracy.

    \item \textbf{DEFT (Training-Free).}
    We run the same base model with our two-turn evaluation prompts:
    it first produces a candidate-blind plan and then selects among the revealed
    trajectories.

    \item \textbf{JEFT $+$ RLVR ($R^{\mathrm{MCQ}}$).}
    GRPO optimizes the single-turn JEFT response using only the binary
    exact-choice reward in Eq.~\ref{eq:rlvr-reward}.
    Its reasoning and selection instructions match the corresponding DEFT turns.
    It uses the same candidate sets,
    training scenes,
    verifier,
    and per-step rollout group size as the other RLVR runs.

    \item \textbf{DEFT $+$ RLVR ($R^{\mathrm{MCQ}}$).}
    This baseline uses the same candidate-blind Turn 1 and candidate-revealed Turn 2 as \textbf{DEFT-RLVR},
    jointly optimizes both generated turns,
    and assigns the same exact-choice reward to their tokens.

    \item \textbf{DEFT $+$ RLVR ($R^{\mathrm{MCQ}}R^{\mathrm{GEN}}$).}
    This variant preserves the two-turn interface and applies a shared
    image-conditioned rubric to outcome-correct rollouts.
    We provide \(\widetilde y_{1,i,j}\) as defined above,
    together with twelve scene frames, to a Qwen3.6-35B-A3B grader.
    The grader is decoded with temperature $1.0$, top-$p=0.7$, a
    $128$-token output limit.
    It returns four binary indicators
    $b^{\mathrm{ground}}$,
    $b^{\mathrm{no\text{-}hall}}$,
    $b^{\mathrm{spec}}$, and
    $b^{\mathrm{coh}}$ for grounding, absence of hallucination, specificity,
    and coherence, respectively.
    We define the shared-rubric score as
    \begin{equation}
    \begin{aligned}
    R^{\mathrm{GEN}}
    ={}&0.30b^{\mathrm{ground}}
    +0.30b^{\mathrm{no\text{-}hall}}
    \\
    &+0.10b^{\mathrm{spec}}
    +0.30b^{\mathrm{coh}},
    \end{aligned}
    \label{eq:generic-rubric-reward}
    \end{equation}
    and assign the rollout reward
    $R=R^{\mathrm{MCQ}}R^{\mathrm{GEN}}$.
    Thus, an invalid or incorrect final choice receives zero before rubric
    grading.
    The rubric, decision rules, and weights are fixed across scenes rather than
    generated per instance; the exact grader prompt is shown in
    Figure~\ref{fig:legacy-judge-template}.

    \item \textbf{JEFT Distillation.}
    We supervise the student with teacher responses produced under JEFT.

    \item \textbf{DEFT Distillation (Plan Only).}
    The student imitates the Qwen3.5-397B-A17B teacher's candidate-blind Turn-1 plan on every annotated scene.
    At evaluation, candidate matching is elicited from the resulting policy without having been included in its SFT targets.

    \item \textbf{DEFT Distillation (Full Interaction).}
    The student imitates both the candidate-blind plan and the subsequent candidate-selection response.

    \item \textbf{DEFT Distillation (Mixed Targets).}
    We divide the same annotated-scene budget approximately equally between Turn-1-only and complete two-turn targets.
\end{itemize}

The three completed deferred-exposure distillation runs use the same $5{,}000$ teacher-labeled scenes,
raw Qwen3-VL-8B-Instruct initialization,
and optimization settings described above.
All principal RLVR variants start from their respective base models;
their shared GRPO configuration is described above.

\paragraph{Runtime Accounting.}
\label{app:runtime-accounting}
Table~\ref{tab:reward-cost} reports median values over deduplicated main-trainer events.
All policy runs use four nodes with $16$ PPU-ZW810E accelerators per node
($96$\,GB per accelerator; $64$ accelerators in total).
The actor and reference model are fully sharded over these $64$ accelerators,
and policy rollouts use vLLM 0.18.0 with $64$ tensor-parallel-size-$1$ engines,
bf16 inference, a $65{,}536$-token context limit,
$0.9$ device-memory utilization, chunked prefill, and CUDA graphs.
The $R^{\mathrm{MCQ}}$-only run uses this policy pool alone.
Both rubric-based variants use an additional, separate pool of
$64$ PPU-ZW810E accelerators for Qwen3.6-35B-A3B grading
(bf16, tensor parallelism $16$).
For $R^{\mathrm{GEN}}$, the online image-conditioned grader receives
the normalized trace \(\widetilde y_{1,i,j}\) and $12$ scene frames; for
\textbf{DEFT-RLVR}, the online grader receives the same normalized trace
without images.
Thus, the table compares end-to-end wall-clock latency under our deployed
configuration, rather than total accelerator-hours.

A step comprises policy rollout, reward scoring, reference-model log
probabilities, the GRPO forward/backward update, and policy-weight
synchronization; validation and pre-training offline rubric generation are
excluded.
The \emph{Rollout} column measures policy generation only, whereas
\emph{Scoring} includes reward queuing, communication, input parsing, and
grader inference.
We aggregate $1{,}234$, $910$, and $1{,}141$ logged steps for
$R^{\mathrm{MCQ}}$, $R^{\mathrm{MCQ}}R^{\mathrm{GEN}}$, and
\textbf{DEFT-RLVR}, respectively, merging resumed logs and retaining the
latest event for each duplicated global step.
No logged warm-up or anomalous steps are manually removed.

\DEFTRLVRTrainingAlgorithm

\subsection{Cold-Start SFT}
\label{app:cold-start-sft}

The cold-start ablation uses $1{,}941$ deduplicated examples:
$1{,}600$ causal-planning examples (an equal mixture of Turn-1-only and complete two-turn targets) and
$341$ general multimodal reasoning examples relabeled by the teacher.
We hold the preprocessing seed fixed at $20260625$.
This SFT stage freezes the visual encoder and updates the language model without resizing the original $151{,}936$-entry vocabulary.
Optimization uses a constant learning rate of $1\!\times\!10^{-5}$ with
$10$ warmup steps, a global batch size of $64$, a per-rank micro batch size
of $1$, a maximum sequence length of $24{,}576$, and tensor parallelism of
$2$. The main run trains for five epochs, comprising $155$ optimizer steps
($31$ per epoch), and saves every $31$ steps. For the dense initialization
study, we use the same configuration for one epoch and save every $3$ steps.

\subsection{Direct Trajectory-Token SFT Diagnostic}
\label{app:interface-diagnostic}
\label{app:direct-token-sft}

This controlled experiment underlies Figure~\ref{fig:traj-overfit} and separates
three possible bottlenecks:
trajectory quantization,
inference over the trajectory-token vocabulary,
and retention of the base VLM's general capabilities.
It is distinct from the adaptation setting in Section~\ref{sec:exp-setup} and
uses its own matched SFT configuration.

\paragraph{Model, Data, and Targets.}
We extend Qwen3-VL-8B-Instruct with the $K{=}8192$ trajectory tokens analyzed
in Appendix~\ref{app:traj-token-repr}.
Each token decodes to a $5\,$s ego-frame trajectory containing 10 waypoints at
$2\,$Hz.
Both settings use the same $100{,}000$ examples:
$88{,}636$ Waymo-E2E training scenes and $11{,}364$ internal driving scenes.
\emph{SFT wo/ CoT} directly emits the oracle trajectory token.
\emph{SFT w/ CoT} first emits a four-part rationale---scene description,
critical object,
reasoning,
and best action---annotated by Qwen3.5-397B-A17B with the GT action available,
and then emits the same oracle token.
Accordingly,
this setting tests trajectory-anchored rationalization rather than the scene-first
reasoning used by \textbf{DEFT-RLVR}.

The teacher annotation limit is $2048$ tokens.
Of the raw CoT annotations,
$6{,}796$ ($6.8\%$) end before the requested ``Best Driving Action'' conclusion
and $16{,}104$ ($16.1\%$) lack a parseable structured action field.
Preprocessing repairs the output wrapper so that every final training target
contains a closed reasoning segment and answer segment;
the underlying truncation remains a limitation of this diagnostic.

\paragraph{Optimization.}
We train both settings for 14 epochs with global batch size $128$,
micro-batch size $1$,
maximum sequence length $2048$,
a constant learning rate of $1\!\times\!10^{-5}$,
and $100$ warmup steps.
One epoch corresponds to 782 optimizer steps.
All optimization and data settings other than the target sequence are shared.

\paragraph{Trajectory Evaluation.}
We distinguish memorization,
held-out in-domain generalization,
and zero-shot cross-domain transfer.
The memorization tier uses seen internal moving scenes;
the in-domain tier uses Waymo-E2E validation scenes excluded from training;
and the OOD tier uses nuScenes, which is absent from the training mixture.
We decode each predicted special token without skipping special tokens and
report ADE, FDE, and parse-failure rate over the full $5\,$s horizon.
The epoch curves use fixed $n{=}100$ subsets.
At epoch~14,
larger $n{=}2000$ evaluations give $2.136\,$m ADE for NoCoT and $2.257\,$m
for CoT on Waymo-E2E validation;
the available NoCoT nuScenes endpoint is $5.297\,$m.
These larger endpoints agree with the trends in
Figure~\ref{fig:traj-overfit}a--b.

\paragraph{General-Capability Evaluation.}
We use greedy decoding with one sample,
no CoT prompt,
and model thinking disabled.
We evaluate CV-Bench-2D/3D,
DA-2K,
ERQA,
EmbSpatialBench,
RoboSpatialHome,
MMSIBench,
RefSpatial-Bench Location/Placement/Unseen,
3DSRBench,
and ViewSpatialBench,
and report the arithmetic mean of their 12 primary metrics as AVG(12).

Although training loss continues to decrease,
held-out Waymo-E2E ADE reaches its minimum near epoch~8 and then rises,
while zero-shot nuScenes ADE plateaus after epoch~4.
The best held-out ADE is $1.963\,$m,
compared with the subset-specific $0.279\,$m codebook quantization floor
($0.290\,$m on the larger held-out representation split in Appendix~\ref{app:traj-token-repr}).
Adding CoT is slightly worse in domain and provides only a small out-of-domain buffer,
with additional parse failures.
The two settings therefore reach similar trajectory accuracy despite substantially different general-capability retention.

For SFT wo/ CoT,
the $12$-benchmark mean falls from $53.49\%$ at epoch~1 to $0.67\%$ at epoch~8 and approaches zero thereafter.
CoT slows but does not prevent forgetting:
its mean decreases from $50.93\%$ to $21.50\%$ by epoch~14.
These curves show that rationale supervision primarily delays destructive specialization rather than improving trajectory precision.

We additionally compare the output interfaces directly using Qwen3.5-397B-A17B on the hard-causal development set.
Full-vocabulary prediction obtains approximately zero strict accuracy and $2.5\,$m ADE,
whereas six-way candidate selection reaches $0.61$ greedy accuracy and $0.88$ pass@$64$.
This is not a matched downstream evaluation,
but it isolates the output interface and supports the conclusion that candidate restriction substantially reduces search difficulty.

\subsection{Candidate-Set Difficulty and Construction Robustness}
\label{app:candidate-set-ablation}

This diagnostic tests whether the advantage of \textbf{DEFT-RLVR} persists when candidate-set difficulty changes.
We hold the scene,
question,
and oracle trajectory fixed and rebuild only the distractors.
Unlike the structured Dev/Test construction in Appendix~\ref{app:ad-mcq-construction},
which combines scale-matched,
constant-velocity,
and hard-negative candidates,
this diagnostic disables the first two sources and samples every distractor from the hard-negative similarity band.
It therefore isolates candidate count and band width rather than reproducing the main \textbf{AD-MCQ-500} candidate sets.

We vary the number of candidates as $M\in\{2,4,6,8,10\}$ and the upper similarity bound as
$\rho_{\max}\in\{0.50,0.70,0.85,0.95\}$,
while fixing $\rho_{\min}=0.30$.
Similarity is derived from the pairwise trajectory ADE within the codebook:
$\rho=1-(\mathrm{ADE}-d_{\min})/(d_{\max}-d_{\min})$.
For each setting,
we evaluate \textbf{DEFT (Training-Free)},
the \textbf{DEFT-RLVR} checkpoint,
and \textbf{DEFT Distillation (Mixed Targets)} with two candidate-blind rounds
of eight samples per question.
The adapted checkpoints are \textbf{DEFT-RLVR} step $570$ and
\textbf{DEFT Distillation (Mixed Targets)} iteration $5750$;
decoding follows the AD-specific evaluation configuration in Appendix~\ref{app:evaluation-settings}.
One malformed item is excluded consistently,
leaving $499$ paired questions.

\begin{table}[h]
\centering
\begin{tabular}{ccllll}
\toprule
$M$ & $\rho_{\max}$ & \textbf{DEFT-TF} & \textbf{DEFT-RLVR} & \textbf{Mixed Distill.} & $\boldsymbol{\Delta}$ \\
\midrule
2 & 0.50 & 88.3 & 95.8 & 97.6 & $+7.5$ \\
2 & 0.70 & 88.1 & 95.2 & 96.9 & $+7.1$ \\
2 & 0.85 & 84.8 & 90.8 & 94.5 & $+6.0$ \\
2 & 0.95 & 75.3 & 79.7 & 90.5 & $+4.4$ \\
\midrule
4 & 0.50 & 73.3 & 88.7 & 98.8 & $+15.4$ \\
4 & 0.70 & 73.5 & 89.2 & 98.3 & $+15.7$ \\
4 & 0.85 & 67.8 & 83.4 & 95.0 & $+15.6$ \\
4 & 0.95 & 51.3 & 61.0 & 84.7 & $+9.7$ \\
\midrule
6 & 0.50 & 65.6 & 85.7 & 99.0 & $+20.0$ \\
6 & 0.70 & 71.3 & 88.4 & 97.7 & $+17.0$ \\
6 & 0.85 & 67.3 & 81.0 & 94.3 & $+13.7$ \\
6 & 0.95 & 45.8 & 55.6 & 81.2 & $+9.8$ \\
\midrule
8 & 0.50 & 57.3 & 82.1 & 98.7 & $+24.8$ \\
8 & 0.70 & 68.3 & 85.5 & 98.3 & $+17.2$ \\
8 & 0.85 & 60.6 & 78.1 & 93.7 & $+17.5$ \\
8 & 0.95 & 39.2 & 49.5 & 77.8 & $+10.3$ \\
\midrule
10 & 0.50 & 44.3 & 64.1 & 79.5 & $+19.8$ \\
10 & 0.70 & 51.3 & 67.4 & 78.0 & $+16.1$ \\
10 & 0.85 & 47.0 & 60.7 & 73.3 & $+13.7$ \\
10 & 0.95 & 29.8 & 36.0 & 59.7 & $+6.2$ \\
\bottomrule
\end{tabular}
\caption{Candidate-set ablation accuracy (\%).
$\Delta$: \textbf{DEFT-RLVR} gain over \textbf{DEFT (Training-Free)}.
Pure hard-negative distractors are resampled; each entry averages two
candidate-blind rounds (eight samples per question).}
\label{tab:candidate-set-ablation}
\end{table}
\textbf{DEFT-RLVR improves over DEFT (Training-Free) in all $20$ settings,
with gains ranging from $4.4\%$ to $24.8\%$.}
The two-candidate setting places both models near a ceiling and offers limited discrimination.
At $\rho_{\max}=0.95$,
both models are compressed by closely matched distractors,
and the gain also narrows.
Even at $M{=}10$ and $\rho_{\max}{=}0.95$,
both models remain above the $10\%$ chance level,
so the hardest setting remains discriminative rather than collapsing to random choice.
Intermediate thresholds exhibit small non-monotonic variation,
so we do not interpret $\rho_{\max}$ as a perfectly calibrated scalar measure of realized difficulty.
\textbf{DEFT Distillation (Mixed Targets)} attains the highest accuracy throughout this grid,
but its training targets and objective differ from those of \textbf{DEFT-RLVR};
the comparison is therefore descriptive rather than a controlled RL-versus-SFT attribution.

\section{Human Validation of AD CoT Evaluation}
\label{app:human-judge-validation}

We validate the automatic evaluation used for the two AD CoT metrics in the
main results.
We first pool candidate-blind Turn-1 CoTs from all methods evaluated on
\textbf{AD-MCQ-500} and then randomly sample $200$ outputs from this combined
pool.
This audit is designed to measure human--judge agreement rather than compare
individual methods,
so the sample is not stratified by method.
We randomly partition the sampled outputs into two disjoint subsets and assign
one subset to each of two human annotators.
The annotators work independently on their assigned subsets without seeing the
model identity or the scores produced by the automatic judge,
and each output receives exactly one human annotation.
For CFS,
the human annotators use the four-dimension $0/1/2$ ordinal rubric defined in
Appendix~\ref{app:motivation-study},
summing grounding, absence of hallucination, specificity, and causal coherence
to a score in $[0,8]$ and then normalizing it to $[0,1]$.
The automatic judge evaluates the same four conceptual dimensions but makes
the binary decisions defined in Appendix~\ref{app:open-eval-judge-prompts};
its four $0/1$ outputs are averaged to obtain the automatic Normed-CFS in
$[0,1]$.
We compute CFS agreement after placing both scores on this common normalized
scale.
For HLD,
both the human annotator and automatic judge require the predicted direction
and speed decision to agree with the action induced by the GT trajectory.
We pool the resulting $200$ non-overlapping human--judge pairs to compute the
agreement statistics in Table~\ref{tab:human-judge-validation}.
The Qwen3.5-397B-A17B judge and its prompt were fixed before the human labels
were examined.

\begin{table}[ht]
\centering
\begin{tabular}{llc}
\toprule
\textbf{Metric} & \textbf{Agreement Measure} & \textbf{Result} \\
\midrule
CFS & Spearman's $\rho$ & $0.78$ \\
CFS & Mean absolute error & $0.067$ \\
CFS & Within $0.125$ agreement & $88.5\%$ \\
HLD & Exact agreement & $92.0\%$ \\
HLD & Cohen's $\kappa$ & $0.84$ \\
\bottomrule
\end{tabular}
\caption{Agreement between human annotations and the automatic judge on a
random sample of $200$ candidate-blind AD reasoning outputs.
The two annotators evaluate disjoint subsets, so each output contributes one
human--judge pair.
The $0.125$ tolerance corresponds to one point on the human annotator's
unnormalized $0$--$8$ CFS scale.}
\label{tab:human-judge-validation}
\end{table}

Across both metrics, the human annotations are highly consistent with the
automatic scores: CFS exhibits strong rank agreement and small absolute error,
while HLD decisions show high exact and chance-corrected agreement.
This audit supports the use of the automatic judge for the AD CoT metrics in
the main table; candidate-selection accuracy remains exact-match based and is
therefore outside the scope of this validation.

\section{Prompt Templates}
\label{app:prompt-templates}

\definecolor{promptblue}{HTML}{0070C0}
\definecolor{promptlightblue}{HTML}{4A90D9}
\definecolor{promptbg}{HTML}{F7F7F7}
\definecolor{promptborder}{HTML}{E8E8E8}
\definecolor{promptcyan}{HTML}{00838F}
\definecolor{promptorange}{HTML}{C65F00}
\definecolor{promptgreen}{HTML}{16833B}
\definecolor{promptpurple}{HTML}{7A245C}
\definecolor{jeftpink}{RGB}{231,161,176}

\newcommand{\prompttitle}[1]{%
  {\color{promptblue}\bfseries #1}\par\vspace{2pt}}
\newcommand{\promptlabel}[1]{%
  {\color{promptblue}\bfseries #1}}
\newcommand{\promptvar}[1]{\texttt{\char`\{#1\char`\}}}
\newcommand{\promptdivider}{%
  \par\medskip{\color{promptborder}\hrule height 0.5pt}\medskip}
\newcommand{\promptdividercompact}{%
  \par\vspace{2pt}{\color{promptborder}\hrule height 0.5pt}\vspace{2pt}}
\ifdefined\DEFTChineseTranslation
\newtcolorbox{deftpromptbox}[2][]{%
  breakable,
  colback=promptblue!3!white,
  colframe=promptblue!75!black,
  title={#2},
  fonttitle=\bfseries\small,
  fontupper=\footnotesize,
  boxrule=0.8pt,
  arc=1.5mm,
  left=1.5mm,
  right=1.5mm,
  top=1.2mm,
  bottom=1.2mm,
  before skip=5pt,
  after skip=4pt,
  #1
}
\else
\newtcolorbox{deftpromptbox}[2][]{%
  colback=promptblue!3!white,
  colframe=promptblue!75!black,
  title={#2},
  fonttitle=\bfseries\small,
  fontupper=\footnotesize,
  boxrule=0.8pt,
  arc=1.5mm,
  left=1.5mm,
  right=1.5mm,
  top=1.2mm,
  bottom=1.2mm,
  before skip=5pt,
  after skip=4pt,
  #1
}
\fi
\newcommand{\deftfigurecaption}[2]{%
  \par\vspace{3pt}
  \refstepcounter{figure}\label{#1}%
  \noindent\hfill\begin{minipage}{0.96\linewidth}
  \small\textbf{Figure~\thefigure:} #2
  \end{minipage}\hfill\mbox{}\par\vspace{6pt}
}
\newcommand{\DEFTPolicyTurnOnePromptTemplate}{%
\begin{deftpromptbox}[fontupper=\footnotesize,
  top=0.8mm,
  bottom=0.8mm,
  before skip=0pt,
  after skip=0pt
]{Turn-1 Candidate-Blind Policy Prompt}
\prompttitle{System Prompt}
You are a helpful assistant.
\promptdividercompact
\prompttitle{User Message --- Turn 1: Plan}
\promptlabel{Visual input:} Video 1 (front-left camera): \texttt{<0.2 seconds><1.2 seconds>}\par
Video 2 (front camera): \texttt{<0.2 seconds><1.2 seconds>}\par
Video 3 (front-right camera): \texttt{<0.2 seconds><1.2 seconds>}\par
\vspace{0.5pt}
Multiple forward-facing camera streams are mounted on the ego vehicle; the frames above are ordered from oldest to newest.\par
\vspace{0.5pt}
The ego vehicle behavior in the recent history is \textbf{\promptvar{EGO\_HISTORY}}. The ego vehicle's current velocity is \promptvar{VEL\_X} m/s at x-direction and \promptvar{VEL\_Y} m/s at y-direction. The ego vehicle's current acceleration is \promptvar{ACC\_X} m/s$^2$ at x-direction and \promptvar{ACC\_Y} m/s$^2$ at y-direction. The current driving command instruction of ego vehicle is: \promptvar{DRIVING\_CMD}, indicating the intended route direction. Note that the left and right driving commands cover turns, lane changes and sharp curves driving behavior.\par
\vspace{0.5pt}
Trajectory coordinates are in meters in the ego frame: +x is forward, +y is to the left. Plan the ego's 5-second future motion (t = 0.5s .. 5.0s).\par
\vspace{0.5pt}
This is a multiple-choice problem, but solve it in TWO option-independent parts. In Part 1 reason CAUSALLY, as if you must plan the ego trajectory yourself; ignore that options exist until Part 2. Use only history-visible evidence; never cite unseen future events.\par
\vspace{0.5pt}
\promptlabel{PART 1 --- PLAN (no options are shown to you; plan the ego maneuver from the scene alone)}\par
In 1--2 concise sentences each:\par
\textbf{1. Evidence:} from the history-visible scene, identify only the few facts that causally constrain the future motion --- the road/route structure and any element whose state changes what the ego can safely, legally, or feasibly do next.\par
\textbf{2. Causal chain:} for each constraint reason scene $\rightarrow$ consequence (what becomes unsafe, illegal, infeasible, or off-route) $\rightarrow$ the maneuver it forces. Treat the ego's CURRENT motion (its speed and acceleration) as only the starting condition to be acted upon, NEVER as evidence that the motion should continue unchanged: the plan follows from the scene, not from the current velocity. Unless the scene positively shows that continuing unchanged is safe and on-route, let the constraints --- not the momentum --- decide the maneuver.\par
\textbf{3. SPEED first, and firmly:} decide whether to keep speed, slow, or stop from the hazards alone (signals, a lead or stopped vehicle, a crossing agent, a stop line, a tight curve or intersection). This decision holds REGARDLESS of which way the road goes --- never default to holding speed just because the path looks clear or because you are unsure where the road leads.\par
Then commit to a high-level decision on its own line in EXACTLY this format:\par
\texttt{HIGH\_LEVEL\_DECISION: <one firmly committed speed profile (e.g. decelerate to a stop / hold $\sim$Xm/s / slow then proceed) + exactly one firmly committed direction from straight / left / right; one sentence, with the main cause>}
\end{deftpromptbox}
\deftfigurecaption{fig:policy-turn1-template}{Turn-1 policy chat template. The policy observes the scene, ego state, and navigation command, but no candidate trajectory. The role-separated panel preserves the production message order, and variables in braces are instantiated per scene.}
}

\newcommand{\DEFTPolicyTurnTwoPromptTemplate}{%
\begin{deftpromptbox}[colback=promptcyan!3!white,
  colframe=promptcyan!75!black,
  fontupper=\footnotesize,
  top=0.8mm,
  bottom=0.8mm,
  before skip=0pt,
  after skip=0pt
]{Turn-2 Candidate-Trajectory Matching Prompt}
\prompttitle{System Prompt}
Continuation of Turn 1; no new system message.
\promptdividercompact
\prompttitle{User Message --- Turn 2: Option Match}
\promptlabel{PART 2 --- OPTION MATCH (only now use the options)}\par
Below are the candidate future trajectories. Exactly one is the best choice.\par
\vspace{0.5pt}
\texttt{Option A}\par
\quad\texttt{waypoints xy: [(x0,y0), ... 10 points ...]}\par
\texttt{Option B}\par
\quad\texttt{waypoints xy: [...]}\par
\texttt{Option C}\par
\quad\texttt{waypoints xy: [...]}\par
\texttt{Option D}\par
\quad\texttt{waypoints xy: [...]}\par
\texttt{Option E}\par
\quad\texttt{waypoints xy: [...]}\par
\texttt{Option F}\par
\quad\texttt{waypoints xy: [...]}\par
\vspace{0.5pt}
Pick the candidate that best realizes your \texttt{HIGH\_LEVEL\_DECISION}. Both the committed SPEED profile (keep speed / slow / stop) and the single committed DIRECTION (straight / left / right) are binding: rule out any candidate inconsistent with either decision. Use the candidate waypoints only to match the already committed plan, never to resolve uncertainty or revise Part 1. If no candidate matches perfectly, choose the closest realization of the fixed speed and direction decision and briefly identify the residual mismatch.\par
End your answer with a final line in EXACTLY this format (nothing after it):\par
\texttt{FINAL\_CHOICE: <one letter from [A, B, C, D, E, F]>}
\end{deftpromptbox}
\deftfigurecaption{fig:policy-turn2-template}{Turn-2 policy chat template. The environment reveals six deterministically shuffled candidate trajectories only after the policy has committed to its first-turn plan. The parser uses the last \texttt{FINAL\_CHOICE} field.}
}

\newcommand{\DEFTRubricGenerationPromptTemplate}{%
\begin{deftpromptbox}[colback=promptorange!3!white,
  colframe=promptorange!80!black,
  fontupper=\fontsize{8}{9}\selectfont,
  top=0.8mm,
  bottom=0.8mm,
  before skip=2pt,
  after skip=2pt
]{Offline Question-Specific Rubric-Generation Prompt}
\prompttitle{System Prompt --- Rubric Generation}
Your task is to generate a self-contained rubric for autonomous-driving causal reasoning. You are given only the history-visible driving scene (multi-view frames + ego state) and the task question. You are NOT given the logged future trajectory, candidate trajectories, or an oracle answer. Generate a set of binary, checkable evaluation criteria describing what a scene-grounded, well-reasoned chain-of-thought (CoT) should contain for this scene. The grader who scores a candidate CoT against your criteria will see only the criterion text and the CoT --- not the frames or any answer-related information. Each criterion must therefore state the relevant scene fact explicitly and be verifiable from the CoT text alone.\par
\vspace{1pt}
\promptlabel{CRITICAL CONSTRAINTS:}\par
\textbf{1. HISTORY-ONLY:} Construct every criterion solely from facts clearly supported by the provided history-visible frames, ego state, and navigation instruction. Never infer an unseen future event, hidden traffic-control state, or hypothetical agent behavior. If a cue is visually ambiguous, either omit it or state the uncertainty explicitly; never turn ambiguity into a definite fact.\par
\textbf{2. PROCESS, NOT OUTCOME:} Evaluate how the CoT identifies visible constraints and reasons about their driving implications. Do NOT prescribe an oracle maneuver, exact future speed, trajectory, waypoint sequence, or option letter. Final maneuver correctness is evaluated separately by an exact outcome verifier. A criterion may require the CoT to explain how a visible cue constrains safe, legal, feasible, or on-route motion, but must not assume access to the future that actually occurred.\par
\textbf{3. SCENE-GROUNDED \& CONCRETE:} Each criterion must reference a concrete cue clearly visible in these frames --- a specific agent, vehicle, pedestrian, lane, traffic light, road feature, navigation instruction, or the ego speed/heading. This is what makes grading mechanical and stable. Generic criteria like ``is specific'' or ``is coherent'' are FORBIDDEN; replace them with a concrete cue check.\par
\textbf{4. CAUSAL RELEVANCE, NOT OBJECT LISTING:} Reward a cue only when the CoT connects it to a driving consequence or constraint. Merely mentioning an object is insufficient. Prefer criteria of the form visible scene fact $\rightarrow$ safety/legal/feasibility/route implication; do not reward exhaustive scene description or unrelated object lists.\par
\textbf{5. REASONING FIDELITY:} Criteria should reward reasoning that is grounded in visible cues, free of invented or unsupported facts, internally consistent, and relevant to the immediate driving decision --- not just fluent or confident language.\par
\textbf{6. ANTI-GAMING:} Include 1--2 criteria that guard against reward-hacking, phrased as GOOD behavior with a POSITIVE weight --- e.g., ``avoids asserting a traffic-light state that is not clearly visible,'' ``avoids treating current momentum as sufficient evidence to continue unchanged,'' ``avoids filler self-praise such as I am confident,'' or ``avoids switching language mid-response.'' These are PRESENT (earn the weight) when the CoT does NOT do the bad thing, and NOT\_PRESENT when it does.\par
\vspace{1pt}
\promptlabel{RUBRIC PRINCIPLES (from OnlineRubrics):} Mutually Exclusive \& Collectively Exhaustive; each criterion Atomic (one idea); Binary (yes/no); Self-contained. 6--10 criteria. Weights are POSITIVE integers 1--10 (NO negative weights --- the paper's synthetic rubrics are positive-only; anti-gaming is phrased as positive good-behavior criteria above).\par
Output ONLY a JSON object, no code fence, no extra prose:\par
\texttt{\char`\{"initial\_reasoning": "brief", "rubrics": [\char`\{"criterion": "text", "weight": int\char`\}, ...]\char`\}}
\promptdividercompact
\prompttitle{User Message --- Rubric Generation}
\promptlabel{Task question (turn-1, the student has NOT seen any options yet):}\par
\texttt{Video 1 (front-left camera): <0.2 seconds><1.2 seconds>}\par
\texttt{Video 2 (front camera): <0.2 seconds><1.2 seconds>}\par
\texttt{Video 3 (front-right camera): <0.2 seconds><1.2 seconds>}\par
\vspace{1pt}
\texttt{<ego-state text and the complete PART 1 prompt in Figure~\ref{fig:policy-turn1-template}>}\par
\vspace{1pt}
Ego state: \texttt{velocity\_norm=}\promptvar{VEL}.\par
\vspace{1pt}
Now generate the rubric criteria for evaluating a candidate CoT about this scene, following the system constraints. Output the JSON object.\par
\vspace{1pt}
\promptlabel{Camera stream 1, oldest to newest.}\par
\texttt{[image: <media\_root>/<token>/4\_resize/00.jpg]}\par
\texttt{[image: <media\_root>/<token>/4\_resize/01.jpg]}\par
\texttt{[image: <media\_root>/<token>/4\_resize/02.jpg]}\par
\texttt{[image: <media\_root>/<token>/4\_resize/03.jpg]}\par
\promptlabel{Camera stream 2, oldest to newest.}\par
\texttt{[image: <media\_root>/<token>/2/00.jpg]}\par
\texttt{... (four frames from stream 2) ...}\par
\promptlabel{Camera stream 3, oldest to newest.}\par
\texttt{[image: <media\_root>/<token>/7/00.jpg]}\par
\texttt{... (four frames from stream 3) ...}
\end{deftpromptbox}
\deftfigurecaption{fig:rubric-generation-template}{Offline question-specific rubric-generation chat template. Criteria are constructed solely from history-visible scene context, without the logged future, candidate trajectories, or an oracle label. Each criterion explicitly encodes a concrete scene constraint and its driving implication so that the image- and answer-blind online grader can check it from the CoT alone. The abbreviated repeated Part-1 block is exactly the text in Figure~\ref{fig:policy-turn1-template}.}
}

\newcommand{\DEFTTextGraderPromptTemplate}{%
\begin{deftpromptbox}[colback=promptgreen!3!white,colframe=promptgreen!75!black]{Online Text-Only Rubric-Grader Prompt}
\prompttitle{System Prompt --- Text-Only Grader}
You are a strict text verifier for autonomous-driving reasoning. You see ONLY the numbered criteria and ONE candidate chain-of-thought (CoT) --- you do NOT see any frames. You are NOT told the correct maneuver and must NOT assume any maneuver is correct. You will evaluate the CoT against the criteria by checking what the CoT TEXT states or omits. For EACH criterion: first state a short objective fact about what the CoT text claims or fails to claim, then derive PRESENT or NOT\_PRESENT mechanically from that fact. Treat any cue the CoT asserts as a textual claim only --- do not assume it is grounded in the real scene, since you cannot see the scene. Never let fluency override what the CoT text actually says.\par
\vspace{3pt}
PRESENT semantics: mark PRESENT if the criterion is satisfied by the CoT text, NOT\_PRESENT otherwise. If a criterion has multiple sub-conditions, NOT\_PRESENT unless ALL are met, EXCEPT ``such as''/``for example''/``including'' lists are illustrative (meeting any one suffices). For anti-gaming criteria phrased as good behavior (e.g., ``avoids inventing X,'' ``avoids filler self-praise''): PRESENT when the CoT does NOT do the bad thing, NOT\_PRESENT when it does.
\promptdivider
\prompttitle{User Message --- Text-Only Grader}
Evaluate the candidate chain-of-thought against EACH numbered criterion below. For each, first state a one-line objective fact, then derive PRESENT or NOT\_PRESENT. Output ONLY a JSON object mapping each criterion number to PRESENT or NOT\_PRESENT. No prose around the JSON, no code fence.\par
\vspace{2pt}
\texttt{1. }\promptvar{criterion text}\texttt{ (weight }\promptvar{w1}\texttt{)}\par
\texttt{2. }\promptvar{criterion text}\texttt{ (weight }\promptvar{w2}\texttt{)}\par
\texttt{... (6--10 criteria) ...}\par
\vspace{2pt}
Output exactly: \texttt{\char`\{"1":"PRESENT","2":"NOT\_PRESENT",...\char`\}}\par
\vspace{3pt}
\texttt{<candidate chain-of-thought>}\par
\promptvar{normalized turn-1 CoT}
\end{deftpromptbox}
\deftfigurecaption{fig:text-grader-template}{Online text-only grader chat template. The grader returns one binary decision per stored criterion; the weighted present rate supplies the process score for an outcome-correct rollout.}
}

\newcommand{\DEFTImageRubricPromptTemplate}{%
\begin{deftpromptbox}[colback=promptpurple!3!white,colframe=promptpurple!80!black]{Image-Conditioned Rubric-Judge Prompt}
\prompttitle{System Prompt --- Image-Conditioned Judge}
You are a strict perception verifier for autonomous-driving reasoning. You see the scene (multi-view frames + ego state) and ONE candidate chain-of-thought. You are NOT told the correct maneuver and must NOT assume any maneuver is correct. For each axis you will first state a short objective fact about the frames, then derive the verdict mechanically from that fact. State only what is clearly visible; if a cue is not clearly visible, treat it as absent. Never let fluency override what the frames show.
\promptdivider
\prompttitle{User Message --- Image-Conditioned Judge}
For each axis output ONE line in EXACTLY this form: \texttt{'<AXIS>: <fact-tag> -> <YES|NO>'}. No other text. Use these fact-tags and rules:\par
\promptlabel{GROUNDING:} fact-tag = \texttt{'cited-visible=all'} if every cited object/agent/road-feature is clearly visible, else \texttt{'cited-visible=some/none'}. Rule: all $\rightarrow$ YES; some/none $\rightarrow$ NO.\par
\promptlabel{NO\_HALLUCINATION:} fact-tag = \texttt{'unsupported-claims=none'} if no claim contradicts/is invented by the frames, else \texttt{'unsupported-claims=any'}. Rule: none $\rightarrow$ YES; any $\rightarrow$ NO.\par
\promptlabel{SPECIFICITY:} fact-tag = \texttt{'concrete-cue-named=yes'} if the reasoning names at least one concrete visible cue by type/position, else \texttt{'no'}. Rule: yes $\rightarrow$ YES; no $\rightarrow$ NO.\par
\promptlabel{COHERENCE:} fact-tag = \texttt{'self-contradictions=none'} if no two statements assert incompatible facts, else \texttt{'any'}. Rule: none $\rightarrow$ YES; any $\rightarrow$ NO.\par
\vspace{2pt}
Output EXACTLY these four lines, nothing else:\par
\texttt{GROUNDING: <tag> -> <YES|NO>}\par
\texttt{NO\_HALLUCINATION: <tag> -> <YES|NO>}\par
\texttt{SPECIFICITY: <tag> -> <YES|NO>}\par
\texttt{COHERENCE: <tag> -> <YES|NO>}\par
\vspace{2pt}
\texttt{<twelve frames from the three camera streams>}\par
\texttt{<candidate chain-of-thought>}\par
\promptvar{normalized turn-1 CoT}
\end{deftpromptbox}
\deftfigurecaption{fig:legacy-judge-template}{Image-conditioned four-axis
rubric-judge template for the controlled variant (axis weights:
$0.30/0.30/0.10/0.30$).}
}

\newcommand{\JEFTPolicyPromptTemplate}{%
\begin{deftpromptbox}[colback=jeftpink!7!white,colframe=jeftpink!80!black]{JEFT Joint-Exposure Policy Prompt}
\prompttitle{System Prompt}
You are a helpful assistant.
\promptdivider
\prompttitle{User Message --- Joint Reasoning and Option Selection}
\promptlabel{Visual input:} Video 1 (front-left camera): \texttt{<0.2 seconds><1.2 seconds>}\par
Video 2 (front camera): \texttt{<0.2 seconds><1.2 seconds>}\par
Video 3 (front-right camera): \texttt{<0.2 seconds><1.2 seconds>}\par
\vspace{2pt}
Multiple forward-facing camera streams are mounted on the ego vehicle; the frames above are ordered from oldest to newest.\par
\vspace{2pt}
The ego vehicle behavior in the recent history is \textbf{\promptvar{EGO\_HISTORY}}. The ego vehicle's current velocity is \promptvar{VEL\_X} m/s at x-direction and \promptvar{VEL\_Y} m/s at y-direction. The ego vehicle's current acceleration is \promptvar{ACC\_X} m/s$^2$ at x-direction and \promptvar{ACC\_Y} m/s$^2$ at y-direction. The current driving command instruction of ego vehicle is: \promptvar{DRIVING\_CMD}, indicating the intended route direction. Note that the left and right driving commands cover turns, lane changes and sharp curves driving behavior.\par
\vspace{2pt}
Trajectory coordinates are in meters in the ego frame: +x is forward, +y is to the left. Plan the ego's 5-second future motion (t = 0.5s .. 5.0s) and select the best candidate trajectory.\par
\vspace{2pt}
\promptlabel{Candidate future trajectories (shown from the outset):}\par
\texttt{Option A}\par
\quad\texttt{waypoints xy: [(x0,y0), ... 10 points ...]}\par
\texttt{Option B}\par
\quad\texttt{waypoints xy: [...]}\par
\texttt{Option C}\par
\quad\texttt{waypoints xy: [...]}\par
\texttt{Option D}\par
\quad\texttt{waypoints xy: [...]}\par
\texttt{Option E}\par
\quad\texttt{waypoints xy: [...]}\par
\texttt{Option F}\par
\quad\texttt{waypoints xy: [...]}\par
\vspace{3pt}
Reason CAUSALLY from the history-visible scene and then match that decision to one candidate. Use only history-visible evidence; never cite unseen future events.\par
\vspace{2pt}
\promptlabel{PART 1 --- PLAN}\par
In 1--2 concise sentences each:\par
\textbf{1. Evidence:} from the history-visible scene, identify only the few facts that causally constrain the future motion --- the road/route structure and any element whose state changes what the ego can safely, legally, or feasibly do next.\par
\textbf{2. Causal chain:} for each constraint reason scene $\rightarrow$ consequence (what becomes unsafe, illegal, infeasible, or off-route) $\rightarrow$ the maneuver it forces. Treat the ego's CURRENT motion (its speed and acceleration) as only the starting condition to be acted upon, NEVER as evidence that the motion should continue unchanged: the plan follows from the scene, not from the current velocity. Unless the scene positively shows that continuing unchanged is safe and on-route, let the constraints --- not the momentum --- decide the maneuver.\par
\textbf{3. SPEED first, and firmly:} decide whether to keep speed, slow, or stop from the hazards alone (signals, a lead or stopped vehicle, a crossing agent, a stop line, a tight curve or intersection). This decision holds REGARDLESS of which way the road goes --- never default to holding speed just because the path looks clear or because you are unsure where the road leads.\par
Then commit to a high-level decision on its own line in EXACTLY this format:\par
\texttt{HIGH\_LEVEL\_DECISION: <one firmly committed speed profile (e.g. decelerate to a stop / hold $\sim$Xm/s / slow then proceed) + exactly one firmly committed direction from straight / left / right; one sentence, with the main cause>}\par
\vspace{3pt}
\promptlabel{PART 2 --- OPTION MATCH}\par
Pick the candidate that best realizes your \texttt{HIGH\_LEVEL\_DECISION}. Both the committed SPEED profile (keep speed / slow / stop) and the single committed DIRECTION (straight / left / right) are binding: rule out any candidate inconsistent with either decision. Use the candidate waypoints only to match the already committed plan, never to resolve uncertainty or revise Part 1. If no candidate matches perfectly, choose the closest realization of the fixed speed and direction decision and briefly identify the residual mismatch.\par
End your answer with a final line in EXACTLY this format (nothing after it):\par
\texttt{FINAL\_CHOICE: <one letter from [A, B, C, D, E, F]>}
\end{deftpromptbox}
\deftfigurecaption{fig:jeft-policy-template}{JEFT joint-exposure policy
template with all candidate trajectories visible before reasoning.}
}

\newcommand{\DEFTOpenEvaluationJudgePromptTemplates}{%
\begin{deftpromptbox}[colback=promptpurple!3!white,
  colframe=promptpurple!80!black,
  fontupper=\footnotesize,
  top=0.7mm,
  bottom=0.7mm,
  before skip=2pt,
  after skip=3pt
]{CFS Evaluation Judge Prompt}
\prompttitle{System Prompt}
You are a strict perception verifier for autonomous-driving reasoning. You see the scene (multi-view frames + ego state) and ONE candidate chain-of-thought. You are NOT told the correct maneuver and must NOT assume any maneuver is correct. For each axis you will first state a short objective fact about the frames, then derive the verdict mechanically from that fact. State only what is clearly visible; if a cue is not clearly visible, treat it as absent. Never let fluency override what the frames show.
\promptdividercompact
\prompttitle{User Message}
For each axis output ONE line in EXACTLY this form:
\texttt{'<AXIS>: <fact-tag> -> <YES|NO>'}.
No other text. Use these fact-tags and rules:\par
\promptlabel{GROUNDING:}
fact-tag = \texttt{'cited-visible=all'} if every cited object/agent/road-feature is clearly visible, else \texttt{'cited-visible=some/none'}. Rule: all $\rightarrow$ YES; some/none $\rightarrow$ NO.\par
\promptlabel{NO\_HALLUCINATION:}
fact-tag = \texttt{'unsupported-claims=none'} if no claim contradicts/is invented by the frames, else \texttt{'unsupported-claims=any'}. Rule: none $\rightarrow$ YES; any $\rightarrow$ NO.\par
\promptlabel{SPECIFICITY:}
fact-tag = \texttt{'concrete-cue-named=yes'} if the reasoning names at least one concrete visible cue by type/position, else \texttt{'no'}. Rule: yes $\rightarrow$ YES; no $\rightarrow$ NO.\par
\promptlabel{COHERENCE:}
fact-tag = \texttt{'self-contradictions=none'} if no two statements assert incompatible facts, else \texttt{'any'}. Rule: none $\rightarrow$ YES; any $\rightarrow$ NO.\par
\vspace{1pt}
Output EXACTLY these four lines, nothing else:\par
\texttt{GROUNDING: <tag> -> <YES|NO>}\par
\texttt{NO\_HALLUCINATION: <tag> -> <YES|NO>}\par
\texttt{SPECIFICITY: <tag> -> <YES|NO>}\par
\texttt{COHERENCE: <tag> -> <YES|NO>}\par
\vspace{1pt}
\texttt{<multi-view scene frames and ego state>}\par
\texttt{<candidate chain-of-thought>}\par
\promptvar{candidate-blind Turn-1 CoT}
\end{deftpromptbox}

\begin{deftpromptbox}[colback=promptcyan!3!white,
  colframe=promptcyan!75!black,
  fontupper=\footnotesize,
  top=0.7mm,
  bottom=0.7mm,
  before skip=3pt,
  after skip=2pt
]{HLD Evaluation Judge Prompt}
\prompttitle{System Prompt}
You are a strict trajectory-matching verifier for autonomous-driving decisions. You are given ONE predicted high-level decision (free text: direction + speed regime + cause) and the GROUND-TRUTH future maneuver (a semantic action label + the gt waypoint xy sequence). Your job is to decide whether the PREDICTED decision matches the GROUND-TRUTH maneuver along two independent axes: DIRECTION and SPEED. For each axis first state a short objective fact, then derive the verdict mechanically. Judge the predicted decision ONLY against the ground truth, not against what would be a safe/legal drive.
\promptdividercompact
\prompttitle{User Message}
\texttt{GROUND-TRUTH maneuver label: }\promptvar{oracle action label}\par
\texttt{GROUND-TRUTH waypoints xy: }\promptvar{first 10 oracle waypoints}\par
\texttt{gt summary: }\promptvar{net displacement, number of points, and mean lateral displacement}\par
\vspace{1pt}
\texttt{PREDICTED high-level decision:}\par
\promptvar{HIGH\_LEVEL\_DECISION text}\par
\vspace{1pt}
For each axis output ONE line in EXACTLY this form:
\texttt{'<AXIS>: <fact-tag> -> <YES|NO>'}.
No other text. Use these fact-tags and rules:\par
\promptlabel{DIRECTION:}
fact-tag = \texttt{'pred-dir=<L|R|S|LC>'} derived from the predicted decision's lateral intent (left=L, right=R, straight/forward=S, lane-change=LC, unsure=U); compare to the gt maneuver's lateral intent (from the label + the sign of gt waypoints' y). Rule: same lateral intent $\rightarrow$ YES; different (e.g., pred straight vs.\ gt left) $\rightarrow$ NO; either unsure $\rightarrow$ NO.\par
\promptlabel{SPEED:}
fact-tag = \texttt{'pred-spd=<stop|decel|constant|accel>'} derived from the predicted speed regime; compare to the gt maneuver's speed regime (from the label; use gt waypoints as confirmation: a gt trajectory that keeps moving far = constant/accel, one that halts = stop, one that shortens = decel). Rule: same regime $\rightarrow$ YES; different (e.g., pred stop vs.\ gt acceleration) $\rightarrow$ NO; unsure $\rightarrow$ NO.\par
\vspace{1pt}
Output EXACTLY these two lines, nothing else:\par
\texttt{DIRECTION\_MATCH: <tag> -> <YES|NO>}\par
\texttt{SPEED\_MATCH: <tag> -> <YES|NO>}
\end{deftpromptbox}
}

\subsection{Two-Turn Candidate-Grounded Policy}
\label{app:two-turn-policy-prompt}

Figure~\ref{fig:policy-turn1-template} gives the first-turn message exactly as presented to the policy, up to example-specific variables. Each of the three video streams is decoded online into four historical frames at $2\,$fps. Qwen3-VL groups adjacent frames into temporal patches, so the rendered message displays two temporal-patch timestamps, $0.2$ and $1.2\,$s; these two markers still correspond to four input frames. Candidate trajectories are deliberately absent from this message. The system message is shared by both turns, and generation uses temperature $1.0$, top-$p$ $0.95$, and a maximum of $12{,}000$ tokens.

\DEFTPolicyTurnOnePromptTemplate


\enlargethispage{4\baselineskip}
\DEFTPolicyTurnTwoPromptTemplate

\clearpage
\subsection{Offline Question-Specific Rubric Generation}
\label{app:offline-rubric-generation-prompt}

For DEFT-RLVR, a fixed Qwen3.6-35B-A3B rubric generator receives twelve scene frames, the first-turn task, and the ego-state summary, but no logged future, candidate trajectory, oracle label, or statistic derived from the future trajectory. The only scalar repeated outside the first-turn task is the current ego-speed norm, computed from the current planar velocity. The generator runs once offline with temperature $0.3$, top-$p$ $0.9$, and a maximum of $1{,}024$ tokens. The resulting six to ten scene-specific criteria are stored with the training example and subsequently applied by the online text-only grader.

\DEFTRubricGenerationPromptTemplate

\subsection{Online Text-Only Rubric Grader}
\label{app:online-text-grader-prompt}

At rollout time, the grader sees only the stored criteria and the normalized
first-turn CoT \(\widetilde y_1\). It receives no image, oracle trajectory,
option list, or high-level-decision line. We use temperature $1.0$, top-$p$
$0.7$, and a maximum of $256$ tokens.

\DEFTTextGraderPromptTemplate

\subsection{Image-Conditioned Rubric Reward for the Controlled Variant}

For \textbf{DEFT $+$ RLVR
($R^{\mathrm{MCQ}}R^{\mathrm{GEN}}$)}, a Qwen3.6-35B-A3B judge scores
\(\widetilde y_1\) from twelve scene frames using temperature $1.0$,
top-$p$ $0.7$, and at most $128$ tokens. This controlled variant retains the
two-turn DEFT interface but replaces \textbf{DEFT-RLVR}'s question-specific
rubric and text-only grader with a shared image-conditioned four-axis rubric.

\DEFTImageRubricPromptTemplate

\subsection{Joint-Exposure Policy Prompt (JEFT)}
\label{app:jeft-policy-prompt}

Figure~\ref{fig:jeft-policy-template} gives JEFT's matched single-turn prompt:
the same scene context, ego state, navigation command, candidates, reasoning
requirements, and outputs as DEFT, but with all six candidates preceding both
reasoning and the high-level decision. JEFT uses temperature $1.0$, top-$p$
$0.95$, and at most $24{,}576$ tokens, matching DEFT's total generation budget.

\JEFTPolicyPromptTemplate

\clearpage
\subsection{CFS and HLD Evaluation-Judge Prompts}
\label{app:open-eval-judge-prompts}

We use the following fixed prompts to evaluate the two open-ended AD metrics
reported in the main results.
The CFS judge is GT-blind and evaluates the candidate-blind Turn-1 CoT against
the visible scene along grounding, absence of hallucination, specificity, and
coherence.
The HLD judge receives the predicted high-level decision and GT maneuver and
requires agreement in both direction and speed.

\DEFTOpenEvaluationJudgePromptTemplates

\clearpage
\section{Detailed Candidate-Trajectory MCQ Case Studies}
\label{app:mcq-case-study}

Figures~\ref{fig:mcq-case-red-light}--\ref{fig:mcq-case-stop-sign} show two
representative \textbf{DEFT-RLVR} rollouts at signal- and
stop-controlled intersections.


\definecolor{casegray}{HTML}{5C5C5C}
\definecolor{caseplan}{HTML}{1F6FB2}
\definecolor{casematch}{HTML}{16833B}

\newtcolorbox{deftcaseoverview}[1]{%
  colback=black!2!white,
  colframe=casegray,
  title={#1},
  fonttitle=\bfseries\small,
  boxrule=0.8pt,
  arc=1.5mm,
  left=1.5mm,
  right=1.5mm,
  top=1.2mm,
  bottom=1.2mm,
  before skip=4pt,
  after skip=5pt
}
\newtcolorbox{deftcasestep}[3]{%
  equal height group=#1,
  colback=#2!4!white,
  colframe=#2!75!black,
  title={#3},
  fonttitle=\bfseries\footnotesize,
  fontupper=\footnotesize,
  boxrule=0.7pt,
  arc=1.5mm,
  left=1.2mm,
  right=1.2mm,
  top=0.9mm,
  bottom=0.9mm,
  before skip=0pt,
  after skip=0pt
}
\newcommand{\mcqtemporalpanel}[1]{%
  \includegraphics[width=\linewidth,height=0.34\textheight,keepaspectratio]{figs/mcq_case_study/#1}}
\newcommand{\mcqoverview}[6]{%
  \begin{deftcaseoverview}{#1}
  \begin{minipage}[t]{0.60\linewidth}
  \vspace{0pt}\centering
  \mcqtemporalpanel{#2}
  \end{minipage}\hfill
  \begin{minipage}[t]{0.365\linewidth}
  \vspace{0pt}\raggedright\footnotesize
  \textbf{Recent motion}\par #3\par\smallskip
  \textbf{Navigation}\par #4\par\smallskip
  \textbf{Current state}\par #5\par\smallskip
  \textbf{Oracle / policy}\par #6\par\medskip
  \textbf{Temporal input}\par
  Rows are front-left, front, and front-right; columns are frames 00--03
  from oldest to newest. The policy receives all four frames from each
  camera, for 12 frames in total.
  \end{minipage}
  \end{deftcaseoverview}
}

\begin{figure*}[!ht]
\centering
\mcqoverview
  {Case 1: Red-Light Stop on a Wet, Constrained Approach}
  {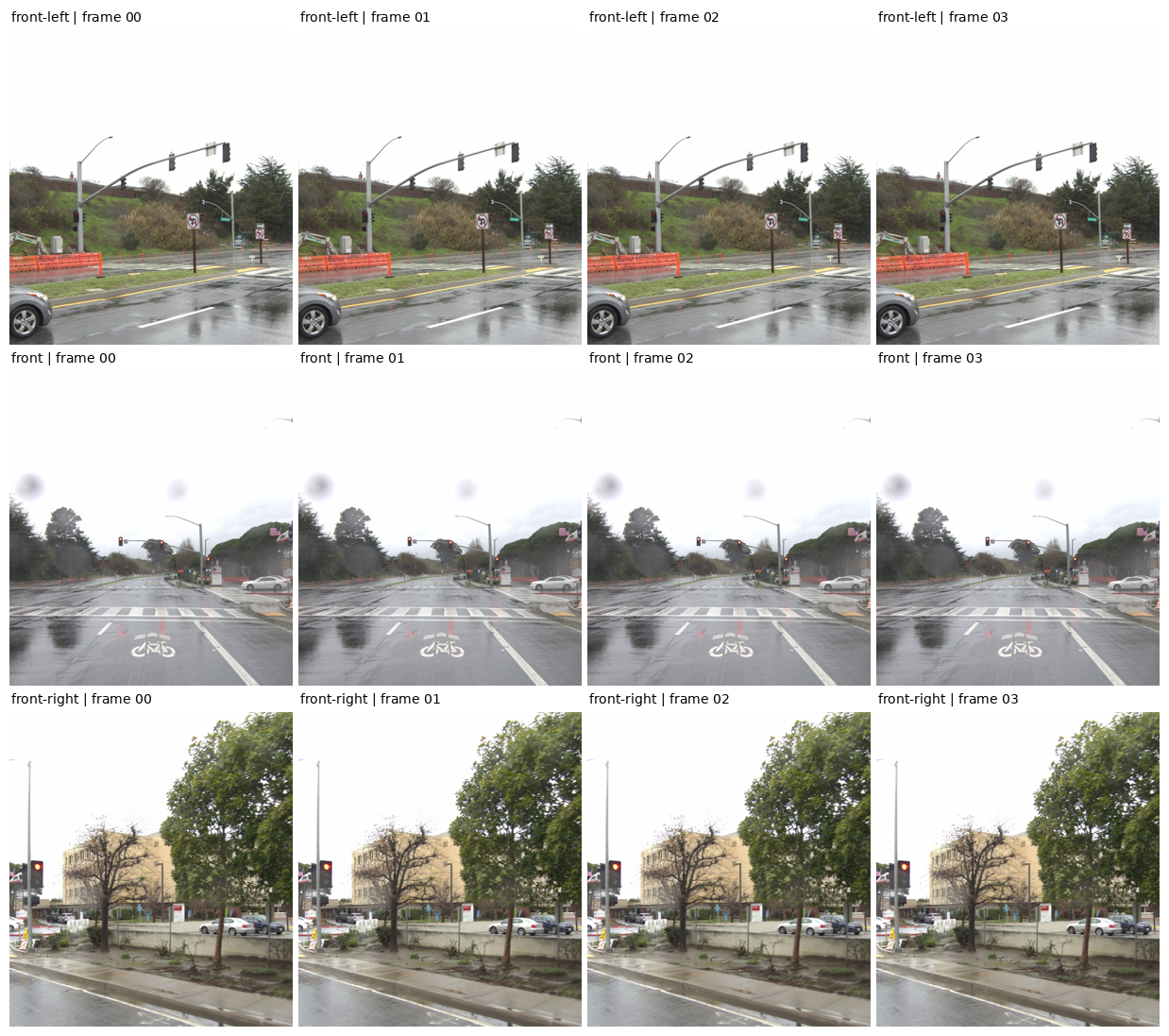}
  {forward with deceleration}
  {go straight}
  {$v_x{=}7.628\,$m/s, $a_x{=}{-}0.595\,$m/s$^2$}
  {A / A}
\noindent
\begin{minipage}[t]{0.485\linewidth}
\vspace{0pt}
\begin{deftcasestep}{mcqcase2}{caseplan}{Turn 1 --- Candidate-Blind Plan}
\textbf{1. Evidence:} The ego is on a wet road approaching a marked
crosswalk with a bike symbol, traffic signals ahead, and construction
barriers on the left. The current deceleration and command to go straight
imply that an intersection or stopping point is imminent.\par
\textbf{2. Causal chain:} The crosswalk and traffic lights require stopping
or slowing to a legal stop before crossing. Continuing at the current speed
risks violating the traffic control or colliding with pedestrians or stopped
vehicles. The construction barriers also restrict lateral maneuvering,
leaving a forward-only path that requires deceleration.\par
\textbf{3. SPEED first:} The ego must decelerate to a stop. The crosswalk and
traffic signals are mandatory stopping cues, and the wet road increases
braking distance, making early deceleration critical.\par
\textbf{4. DIRECTION:} The traffic-control layout, crosswalk markings, and
navigation command constrain the ego to continue straight; no visible cue
supports a turn or curve.\par
\textbf{HIGH\_LEVEL\_DECISION:} decelerate to a stop + direction: straight
ahead---the crosswalk and traffic lights mandate stopping, and the road
geometry does not permit deviation.
\end{deftcasestep}
\end{minipage}\hfill
\begin{minipage}[t]{0.485\linewidth}
\vspace{0pt}
\begin{deftcasestep}{mcqcase2}{casematch}{Turn 2 --- Trajectory-Option Match}
\textbf{A} rapidly reduces its forward waypoint increments, settles near
$x{=}9.9\,$m, and maintains negligible lateral displacement. It therefore
realizes the committed straight stop.\par
\textbf{B--F} all continue moving forward across the horizon. E is slower
than several alternatives but still fails to stop, while B, C, D, and F
accumulate substantially larger longitudinal displacement. These candidates
violate the binding speed decision regardless of their approximate
direction.\par
Only A satisfies both constraints: a stop-like terminal profile and a
straight path with essentially zero lateral deviation.\par
\textbf{FINAL\_CHOICE: A}
\end{deftcasestep}
\end{minipage}
\caption{Complete two-turn MCQ case for a red-light stop on a wet, construction-constrained approach. The policy identifies the signal before seeing options and then matches that commitment to the stopping trajectory A.}
\label{fig:mcq-case-red-light}
\end{figure*}
\clearpage

\begin{figure*}[p]
\centering
\mcqoverview
  {Case 2: Stop-Sign Compliance under a Constant-Speed History}
  {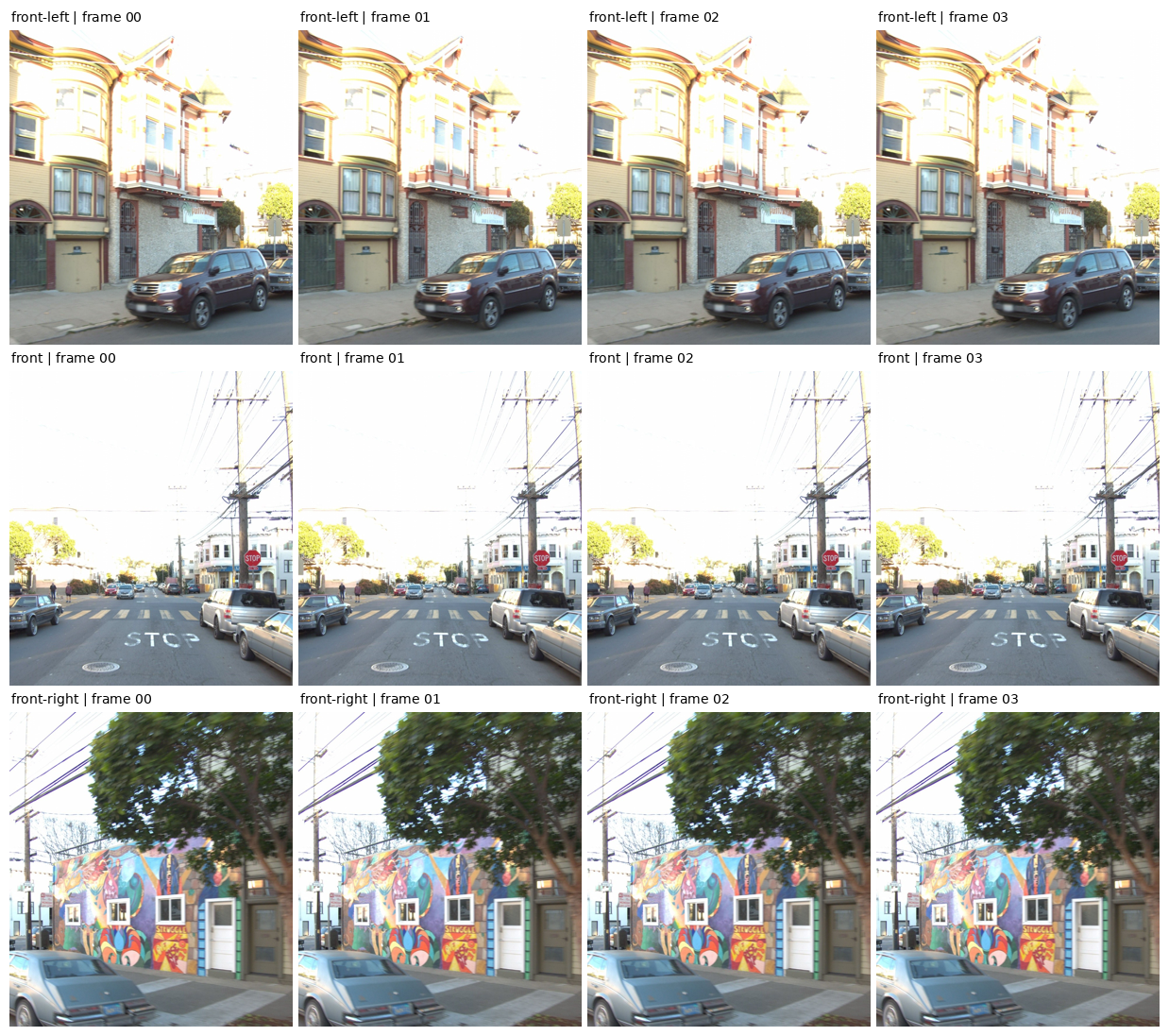}
  {forward at constant speed}
  {go straight}
  {$v_x{=}8.854\,$m/s, $a_x{=}{-}0.455\,$m/s$^2$}
  {C / C}
\noindent
\begin{minipage}[t]{0.485\linewidth}
\vspace{0pt}
\begin{deftcasestep}{mcqcase3}{caseplan}{Turn 1 --- Candidate-Blind Plan}
\textbf{1. Evidence:} The ego approaches a stop line marked \textsc{stop},
with a stop sign visible and vehicles near the approach. The scene is a
street intersection with possible cross-traffic or pedestrian crossings.\par
\textbf{2. Causal chain:} The sign and painted stop line legally mandate
stopping before entry. Continuing from the recent constant-speed history
would violate that control and risk collision with stopped vehicles or
crossing agents. The recent velocity does not override this scene-imposed
constraint.\par
\textbf{3. SPEED first:} The ego must decelerate to zero before proceeding;
the stop sign and stop line force the speed decision.\par
\textbf{4. DIRECTION:} The road appears to continue straight through the
intersection. No visible turning lane or other cue forces a turn, so the
evidence supports proceeding straight once traffic permits.\par
\textbf{HIGH\_LEVEL\_DECISION:} decelerate to a stop + direction: straight
(the stop line and stop sign enforce stopping before proceeding straight).
\end{deftcasestep}
\end{minipage}\hfill
\begin{minipage}[t]{0.485\linewidth}
\vspace{0pt}
\begin{deftcasestep}{mcqcase3}{casematch}{Turn 2 --- Trajectory-Option Match}
\textbf{C} progressively reduces its longitudinal increments to nearly zero
and ends near $(12.86,-0.12)\,$m. Its negligible lateral movement and
stop-like terminal profile match both parts of the prior commitment.\par
\textbf{A, D, and E} continue forward with large longitudinal displacement
and therefore violate the stop requirement. \textbf{B} has substantial
negative lateral displacement, indicating a turn or drift rather than a
straight stop. \textbf{F} traces a clear left turn and likewise violates the
direction constraint.\par
Only C simultaneously realizes the required deceleration to a stop and
negligible lateral deviation.\par
\textbf{FINAL\_CHOICE: C}
\end{deftcasestep}
\end{minipage}
\caption{Complete two-turn MCQ case for a stop-controlled intersection. Despite a constant-speed history, the candidate-blind plan is governed by the visible stop control, and trajectory C is selected only after this commitment.}
\label{fig:mcq-case-stop-sign}
\end{figure*}
\clearpage

\FloatBarrier

\section{Full General Visual Capability Results}
\label{app:general-visual-results}

Tables~\ref{tab:general-visual-basic}--\ref{tab:general-visual-refspatial}
expand the four category aggregates in Table~\ref{tab:main-results} into the
complete 12-benchmark evaluation. JEFT-based methods are task-matched
conventional baselines; DEFT-based rows are variants of our framework.

\begin{table}[htbp]
\centering
\scriptsize
\begin{tabular}{lcccccc}
\toprule
& \multicolumn{3}{c}{\textbf{Basic Visual}} &
\multicolumn{3}{c}{\textbf{Embodied Spatial}} \\
\cmidrule(lr){2-4}\cmidrule(l){5-7}
\textbf{Method} & \textbf{CV2D} & \textbf{CV3D} & \textbf{DA2K}
& \textbf{ERQA} & \textbf{EmbSpat} & \textbf{RoboSpat} \\
\midrule
\textbf{Qwen3-VL-8B-Instruct} & 81.88 & 93.83 & 69.10 & 43.00 & 77.75 & 49.14 \\
\midrule
+ JEFT $+$ RLVR ($R^{\mathrm{MCQ}}$) & 80.97 & 93.92 & 69.29 & 44.00 & 78.30 & 47.71 \\
+ \textbf{DEFT $+$ RLVR ($R^{\mathrm{MCQ}}$)} & 81.41 & 93.58 & 69.10 & 44.75 & 78.41 & 48.29 \\
+ \textbf{DEFT $+$ RLVR ($R^{\mathrm{MCQ}}R^{\mathrm{GEN}}$)} & 81.29 & 94.08 & 69.39 & 42.75 & 78.19 & 48.00 \\
+ \textbf{DEFT-RLVR (ours)} & 80.85 & 94.00 & 69.15 & 45.00 & 77.80 & 49.43 \\
\midrule
+ \textbf{DEFT Distillation (Plan Only)} & 78.12 & 88.67 & 67.21 & 39.25 & 74.09 & 47.14 \\
+ \textbf{DEFT Distillation (Full Interaction)} & 75.96 & 88.50 & 60.35 & 39.00 & 75.69 & 47.43 \\
+ \textbf{DEFT Distillation (Mixed Targets)} & 76.25 & 90.83 & 63.10 & 42.50 & 74.04 & 44.00 \\
\midrule
\textbf{Qwen3.5-4B} & 82.09 & 91.58 & 67.26 & 47.25 & 74.04 & 37.71 \\
\midrule
+ JEFT $+$ RLVR ($R^{\mathrm{MCQ}}$) & 81.55 & 91.58 & 67.55 & 47.75 & 72.83 & 39.43 \\
+ \textbf{DEFT $+$ RLVR ($R^{\mathrm{MCQ}}$)} & 82.27 & 91.50 & 68.52 & 45.75 & 73.65 & 38.86 \\
+ \textbf{DEFT $+$ RLVR ($R^{\mathrm{MCQ}}R^{\mathrm{GEN}}$)} & 81.81 & 92.00 & 67.41 & 47.50 & 73.74 & 38.86 \\
+ \textbf{DEFT-RLVR (ours)} & 82.13 & 92.42 & 68.09 & 50.00 & 73.60 & 42.00 \\
\bottomrule
\end{tabular}%
\caption{Benchmark-level results for \textbf{Basic Visual} and
\textbf{Embodied Spatial} capabilities. Relative to the corresponding base
model, \textbf{DEFT-RLVR} improves five of the six benchmarks for both
backbones, with mean gains of $0.26\%$ for Qwen3-VL-8B and $1.39\%$
for Qwen3.5-4B; the largest gains are $2.00\%$ on ERQA and $4.29\%$
on RoboSpat, respectively.}
\label{tab:general-visual-basic}
\label{tab:general-visual-embodied}
\end{table}

\begin{table}[htbp]
\centering
\scriptsize
\begin{tabular}{lcccccc}
\toprule
& \multicolumn{3}{c}{\textbf{3D/Multi-View}} &
\multicolumn{3}{c}{\textbf{RefSpatial}} \\
\cmidrule(lr){2-4}\cmidrule(l){5-7}
\textbf{Method} & \textbf{3DSR} & \textbf{MMSI} & \textbf{ViewSpat}
& \textbf{RefLoc} & \textbf{RefPlc} & \textbf{RefUns} \\
\midrule
\textbf{Qwen3-VL-8B-Instruct} & 55.21 & 30.70 & 41.54 & 55.00 & 32.00 & 28.57 \\
\midrule
+ JEFT $+$ RLVR ($R^{\mathrm{MCQ}}$) & 54.88 & 29.70 & 41.39 & 57.00 & 37.00 & 24.68 \\
+ \textbf{DEFT $+$ RLVR ($R^{\mathrm{MCQ}}$)} & 54.88 & 31.50 & 42.02 & 56.00 & 43.00 & 33.77 \\
+ \textbf{DEFT $+$ RLVR ($R^{\mathrm{MCQ}}R^{\mathrm{GEN}}$)} & 55.09 & 32.00 & 41.33 & 57.00 & 41.00 & 36.36 \\
+ \textbf{DEFT-RLVR (ours)} & 55.17 & 30.00 & 41.65 & 54.00 & 41.00 & 35.06 \\
\midrule
+ \textbf{DEFT Distillation (Plan Only)} & 50.67 & 27.50 & 42.42 & 45.00 & 32.00 & 23.38 \\
+ \textbf{DEFT Distillation (Full Interaction)} & 49.62 & 27.30 & 44.22 & 44.00 & 35.00 & 24.68 \\
+ \textbf{DEFT Distillation (Mixed Targets)} & 50.42 & 26.80 & 43.38 & 48.00 & 35.00 & 27.27 \\
\midrule
\textbf{Qwen3.5-4B} & 45.98 & 33.50 & 42.72 & 51.00 & 29.00 & 28.57 \\
\midrule
+ JEFT $+$ RLVR ($R^{\mathrm{MCQ}}$) & 45.69 & 32.10 & 42.68 & 44.00 & 24.00 & 25.97 \\
+ \textbf{DEFT $+$ RLVR ($R^{\mathrm{MCQ}}$)} & 48.30 & 32.80 & 44.38 & 49.00 & 36.00 & 31.17 \\
+ \textbf{DEFT $+$ RLVR ($R^{\mathrm{MCQ}}R^{\mathrm{GEN}}$)} & 46.09 & 32.90 & 43.03 & 55.00 & 31.00 & 29.87 \\
+ \textbf{DEFT-RLVR (ours)} & 46.50 & 33.30 & 43.01 & 53.00 & 34.00 & 20.78 \\
\bottomrule
\end{tabular}%
\caption{Benchmark-level results for \textbf{3D/Multi-View} and
\textbf{RefSpatial} capabilities. Relative to the corresponding base model,
\textbf{DEFT-RLVR} changes the Qwen3-VL-8B 3D/Multi-View mean by $-0.21\%$
while improving its RefSpatial mean by $4.83\%$, including gains of $9.00\%$
on RefPlc and $6.49\%$ on RefUns. For Qwen3.5-4B, the six-benchmark mean is
largely preserved ($-0.03\%$), with changes of $+0.20\%$ on 3D/Multi-View
and $-0.26\%$ on RefSpatial.}
\label{tab:general-visual-3dmv}
\label{tab:general-visual-refspatial}
\end{table}

\end{document}